\documentclass[10pt,twocolumn,letterpaper]{article}

\usepackage{iccv}
\usepackage{times}
\usepackage{epsfig}
\usepackage{graphicx}
\usepackage{amsmath}
\usepackage{amssymb}

\usepackage[accsupp]{axessibility}  

\usepackage{booktabs}
\usepackage{flexisym}
\usepackage{multirow}
\usepackage{cite}
\usepackage{subcaption}

\usepackage{tikz}
\usepackage{comment}
\usepackage[misc,geometry]{ifsym}

\usepackage[pagebackref=true,breaklinks=true,letterpaper=true,colorlinks,bookmarks=false]{hyperref}

\iccvfinalcopy 

\def\iccvPaperID{4823} 

\ificcvfinal\fi

\DeclareMathOperator*{\argmin}{arg\,min}

\usepackage{bbm}
\usepackage{color, colortbl}

\begin{document}

\title{Self-Feedback DETR for Temporal Action Detection}
\makeatletter
\def\@fnsymbol#1{\ensuremath{\ifcase#1\or \dagger\or \ddagger\or
\mathsection\or \mathparagraph\or \|\or **\or \dagger\dagger
\or \ddagger\ddagger \else\@ctrerr\fi}}
\makeatother

\author{Jihwan Kim\quad\quad\quad Miso Lee\quad\quad\quad Jae-Pil Heo\thanks{Corresponding author} \\
Sungkyunkwan University \\
{\tt\small \{damien,\;dlalth557,\;jaepilheo\}@skku.edu}
}

\maketitle
\ificcvfinal\thispagestyle{empty}\fi

\begin{abstract}
Temporal Action Detection (TAD) is challenging but fundamental for real-world video applications.
Recently, DETR-based models have been devised for TAD but have not performed well yet.
In this paper, we point out the problem in the self-attention of DETR for TAD; the attention modules focus on a few key elements, called temporal collapse problem.
It degrades the capability of the encoder and decoder since their self-attention modules play no role.
To solve the problem, we propose a novel framework, Self-DETR, which utilizes cross-attention maps of the decoder to reactivate self-attention modules.
We recover the relationship between encoder features by simple matrix multiplication of the cross-attention map and its transpose.
Likewise, we also get the information within decoder queries.
By guiding collapsed self-attention maps with the guidance map calculated, we settle down the temporal collapse of self-attention modules in the encoder and decoder.
Our extensive experiments demonstrate that Self-DETR resolves the temporal collapse problem by keeping high diversity of attention over all layers.
Moreover, it is validated that our simple framework achieves a new state-of-the-art performance on THUMOS14 and outperforms all the DETR-based approaches on ActivityNet-v1.3.
\end{abstract}

\section{Introduction}

\begin{figure}[t]
\centering
\includegraphics[width=8.35cm]{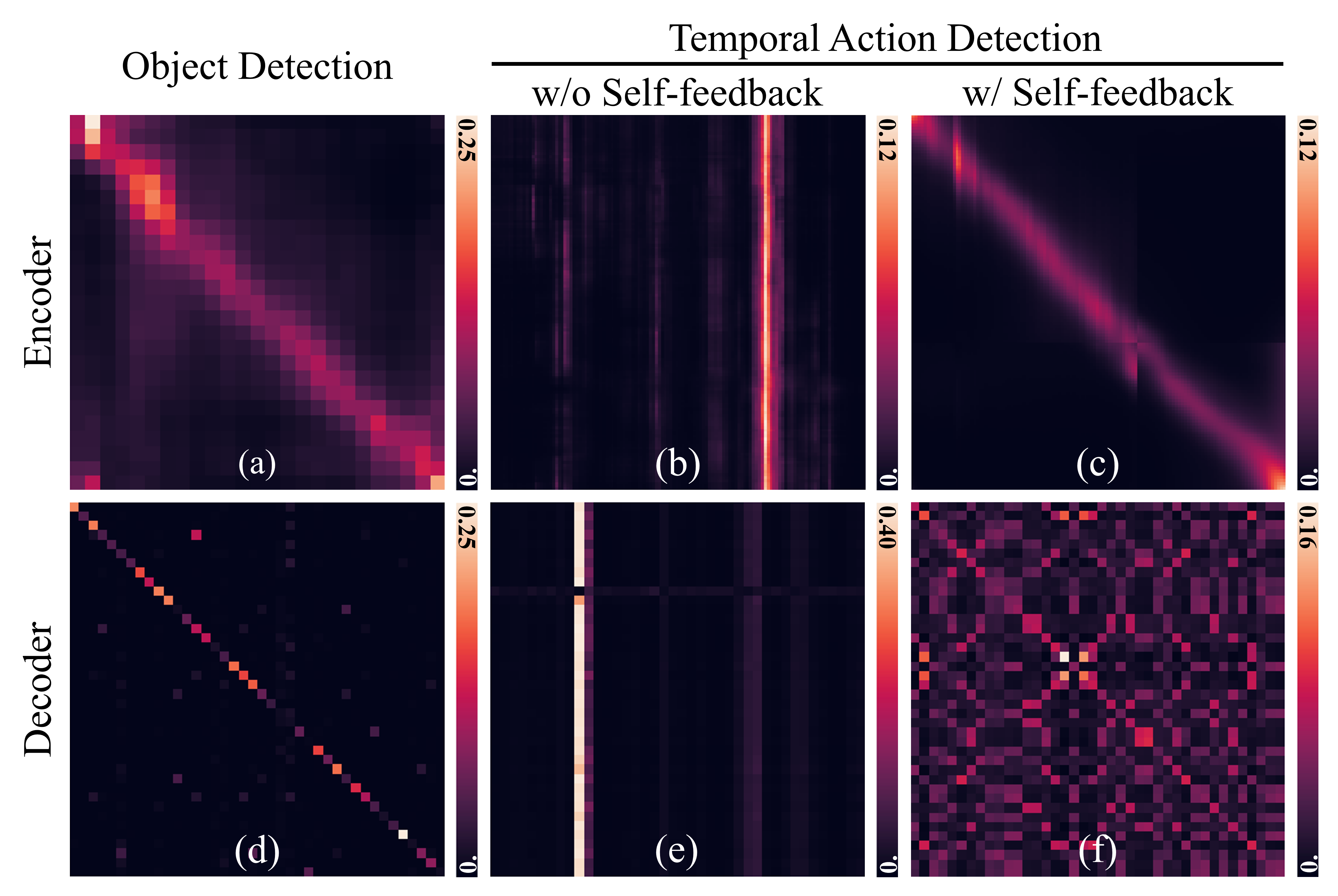}
\vspace{-10pt}
\caption{\textbf{Temporal collapse problem of self-attention.}
The figure shows self-attention maps by DETR-based methods in object detection (OD) and temporal action detection (TAD).
Each map at the top and bottom is from the last layer of the encoder and decoder, respectively.
We can see that the self-attention maps of the encoder and decoder in TAD are collapsed to a small number of keys (b, e).
On the other hand, those from OD and ours show high correlation for neighboring features (a, c) or query themselves (d, f).
}
\label{fig:introduction}
\vspace{-10pt}
\end{figure}

Understanding videos has become fundamental as uncountable videos are produced all over devices every moment.
In the first place, action recognition using trimmed video clips led the field with tremendous advances during past decades.
However, unacceptable costs to snip real-world videos fostered the literature towards temporal action detection (TAD).
Temporal action detection not just classifies an action but also predicts time boundaries of untrimmed video.

Pioneering methods~\cite{escorcia2016dap, caba2016fast, buch2017sst} adopted the concept of fixed-length windows called action proposals inspired by object detection~\cite{girshick2014rcnn, girshick2015fast-rcnn, ren2015faster-rcnn}.
Meanwhile, the following approaches~\cite{zhao2017ssn, yuan2017maximul-sum, lin2018bsn, kim2019coarsefine} developed point-wise learning where they predict probabilities of start and end boundaries to solve the low-recall issue. 
They reached the high-recall score from more flexible action proposals by grouping each pair of start and end boundaries, but unfortunately, a bunch of generated proposals with various lengths made the ranking process more challenging.
Accordingly, the previous methods heavily rely on ranking with post-processing such as non-maximum suppression (NMS) to cope with low-precision action proposals.

As DETR~\cite{carion2020detr} has had a great impact on object detection, DETR-based methods for videos~\cite{tan2021relaxed, liu2021tadtr, shi2022react, lei2021moment-detr, moon2023qd-detr} are also introduced recently.
In TAD, queries of DETR are defined as action instances of a video with their time intervals, called action queries.
Here, the model learns to map these query vectors to relevant temporal features of the video to classify and localize the actions of interest.
Since there is no pre-set mapping between queries and ground-truth instances, bipartite matching associates them with minimal cost for the objectives to assign the labels.
This approach can tackle the task in an end-to-end manner without any heuristics like NMS via the set-based objective.

However, it is discovered that the original DETR architecture suffers from several problems with videos and thereby does not perform well in TAD.
It has been estimated that the main problem is the failure of dense attention mechanism~\cite{tan2021relaxed, liu2021tadtr}.
Dense attention here indicates standard attention mechanism which relates all elements without any inductive bias such as locality in convolution.
To address this issue, previous DETR-based approaches in TAD revised the standard attention to boundary-sensitive module~\cite{tan2021relaxed}, deformable attention~\cite{liu2021tadtr}, or query relational attention~\cite{shi2022react}.

Nevertheless, the problem of the standard attention still remains setting back DETR for TAD far behind in performance.
In this paper, we confront the problem of the standard attention.
Fig.~\ref{fig:introduction} shows self-attention maps of DETR methods.
From the figure, we find that the self-attention for TAD severely suffers from collapse to a few key elements, which we define as \textit{temporal collapse problem}.
This phenomenon implies that the model selects the shortcut to skip the self-attention to elude degeneration of the output.
In contrast, self-attention in object detection and ours is highly correlated without any collapse.
Hence, we point out that the temporal collapse problem in self-attention is the core to degrade DETR-based methods for TAD.

To solve the problem, we propose a new framework, Self-DETR, which provides feedback to self-attention from the encoder-decoder cross-attention.
The cross-attention map contains the entire relation between the encoder and decoder features.
We view the similarity between decoder queries as how much they focus on similar encoder features.
Likewise, we consider the similarity between encoder features as how much they are attended by analogous decoder queries.
We can obtain these two kinds of guidance by simple matrix multiplication of the cross-attention map and its transpose.
From these guidance maps, the temporal collapse is relaxed by minimizing the gap of the guidance and self-attention maps.
Through our extensive experiments on the public benchmarks, 
we validate that Self-DETR settle downs the collapse by retaining high diversity of attention.
As a result, Self-DETR has achieved a new state-of-the-art performance on THUMOS14, and outperforms all the DETR-based methods in ActivityNet-v1.3 without deformable attention.

To sum up, our main contributions are as follows:
\begin{itemize}
	\item We discover the temporal collapse problem of standard attention in the DETR-based models for TAD. We reveal that the main issue is in self-attention of both the encoder and decoder.
	\item We propose a novel framework, Self-DETR, which utilizes cross-attention maps to provide feedback to self-attention of the encoder and decoder to prevent the temporal collapse.
	\item Our extensive experiments demonstrate that Self-DETR blocks the temporal collapse efficiently by keeping high diversity of attention.
    Also we validate that our model reaches a new state-of-the-art performance on THUMOS14, and outperforms all the DETR-based methods on AcitvityNet-v1.3.
\end{itemize}
\section{Related Work}
\subsection{Temporal Action Detection}
Temporal action detection (TAD) is the task to find a time interval of an action instance in an untrimmed video as well as classifying the instance.
Early methods~\cite{yeung2016frame-glimpses, yuan2016score-distribution, escorcia2016dap, shou2016scnn, caba2016fast, buch2017sst, shou2017cdc, yuan2017maximul-sum, xiong2017tag, qiu2018etp, chao2018tal-net, gao2018ctap, kim2019coarsefine} have been realized great improvements in TAD during the last decade. 
As two-stage approaches had been successful in object detection~\cite{girshick2014rcnn, girshick2015fast-rcnn, ren2015faster-rcnn}, a number of methods in temporal action localization deployed multi-stage strategies~\cite{gao2017turn, zhao2017ssn, xu2017rc3d, dai2017tcn, qiu2018etp}.
As another stream of research, R-C3D~\cite{xu2017rc3d} adopted the R-CNN architecture~\cite{ren2015faster-rcnn} in object detection with C3D model~\cite{tran2015c3d} for the first time.
Similarly, TAL-Net~\cite{chao2018tal-net} customized the Faster R-CNN architecture for TAD with dilated convolutions for Region-of-Interest (RoI) pooling.

As the subsequent work, point-wise learning has been widely introduced to generate more flexible action proposals without pre-defined temporal windows.
SSN~\cite{zhao2017ssn} and TCN~\cite{dai2017tcn} extended temporal context around a generated proposal to improve ranking performance.
BSN~\cite{lin2018bsn} and BMN~\cite{lin2019bmn} grouped candidate start-end pairs to generate action proposals, then ranked them for final detection outputs.
BSN++~\cite{su2021bsn++} tackled scale imbalance problem based on BSN.
Besides, graph neural networks are getting a great deal of attention in TAD~\cite{xu2020gtad, zeng2019pgcn}.
PGCN~\cite{zeng2019pgcn} improved ranking performance via constructing a graph of proposals based on their overlaps. 
GTAD~\cite{xu2020gtad} considered TAD as sub-graph localization problem and proposed a new framework with graph neural network.
Also, TCANet~\cite{qing2021temporal} devised local and global temporal context aggregation. ActionFormer~\cite{zhang2022actionformer} deployed transformer encoder as backbone network, and E2E-TAD~\cite{liu2022empirical} studied for the end-to-end learning in TAD.
AMNet~\cite{kang2023amnet} introduced a new framework to refine video features via action-aware attention.

\subsection{DETR}
End-to-end object DEtection with TRansformers (DETR)~\cite{carion2020detr} firstly viewed object detection as a direct set prediction problem, and removed the need of the heuristic process, non-maximum-suppression (NMS).
However, DETR required 10 times longer training time than the conventional approaches since Hungarian matching is hard to be optimized with dense attention.
To cope with this issue, Deformable DETR~\cite{xizhou2021deformable_detr} introduced sparse attention, which attends only a part of elements by learning to specify positions to focus on.
Deformable attention gives locality back to DETR so that the training time is significantly reduced with performance improvement.
The following DETR-based models~\cite{meng2021conditional_detr, shilong2022dab_detr} further improved query representations through explicitly encoding box information, which effectively helps to stabilize training.

Transformer-based models inherently suffer from the large computational cost due to dense attention.
In order to further reduce computational cost, Sparse DETR~\cite{byungseok2022sparse_detr} introduced learnable sparsity to encoder features.
To this end, they utilized encoder-decoder cross-attention maps to produce a binary mask as the guidance of sparsity.
This way is quite related to ours in that they teach to sparsify encoder features by cross-attention relation as features with more attention from the decoder are more likely crucial for the task.

\begingroup
\setlength{\tabcolsep}{6.0pt} 
\renewcommand{\arraystretch}{1.10} 
\begin{table}[t]
	\centering
	\begin{tabular}{l|ccc}
		\hline\hline
		Method & Enc. SA & Dec. SA & Dec. CA \\
		\hline\hline
        RTD-Net~\cite{tan2021relaxed} & MLP & Standard & Standard \\
        TadTR~\cite{liu2021tadtr} & Deformable & Standard & Deformable \\
        ReAct~\cite{shi2022react} & Deformable & Heuristic & Deformable \\
        \hline
        Self-DETR & Standard & Standard & Standard \\
        \hline\hline
	\end{tabular}
    \vspace{-10pt}
	\caption{\textbf{Comparison of baselines and Self-DETR in terms of attention methods}. 
 The table shows the comparison in terms of the methods for self-attention and cross-attention. `Enc.' and `Dec.' mean the encoder and decoder, `SA' and `CA' indicate self-attention and cross-attention.
 }
	\label{tab:related_work}
    \vspace{-15pt}
\end{table}
\endgroup

In TAD, DETR is also introduced recently as DETR-based models have reached a new state-of-the-art performance in object detection.
RTD-Net~\cite{tan2021relaxed} pointed out the problem of the dense attention in the DETR's encoder, which shows nearly uniform distribution so that the self-attention layers act like over-smoothing effect.
RTD-Net replaced the transformer encoder with the boundary-sensitive module to relieve the smoothing effect.
On the other hand, TadTR~\cite{liu2021tadtr} devised temporal deformable attention inspired by Deformable DETR~\cite{xizhou2021deformable_detr} for the same problem of RTD-Net.
When it comes to query relation in the decoder, ReAct~\cite{shi2022react} developed a new relation matching to enforce high correlation between queries with low-overlap and high feature similarity.
This way alleviated the problem of query collapse in self-attention of the decoder since the pre-arranged relations are only permitted.

However, the problem still remains since they have detoured standard attention by deformable attention or heuristic query relation, as compared in Tab.~\ref{tab:related_work}.
However, we directly settle down the problem of self-attention in TAD without any deformable attention or heuristic relation.

To be specific about the problem definition, we clearly identify the problem of self-attention as temporal collapse beyond existing over-smoothing effect.
It is already demonstrated that collapsed rank-1 matrix degrades performances in transformer architecture~\cite{dong2021rank_collapse}.
Here, we find that current situations are fully aligned as shown in Fig.~\ref{fig:introduction}.
Therefore, we claim that the over-smooth matrix is one kind of collapsed matrix but with relatively small values.
\begin{figure*}[t]
\centering
\includegraphics[width=17.40cm]{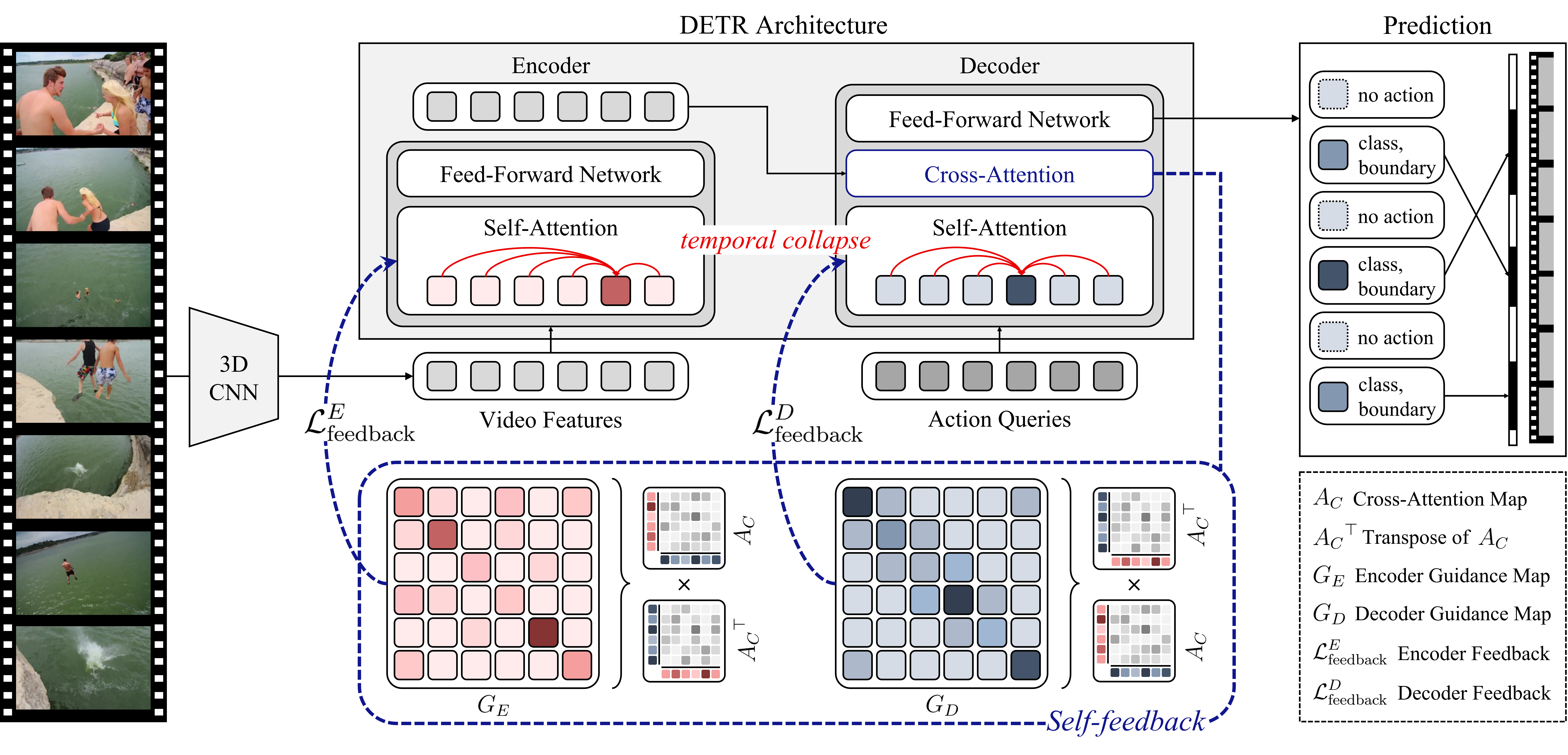}
\vspace{-20pt}
\caption{\textbf{Overall architecture of the proposed framework, Self-DETR.}
Self-DETR is based on DETR architecture, and on top of it, we design self-feedback in order to alleviate the temporal collapse problem in the self-attention modules.
We produce two types of guidance $G_E$ and $G_D$ for the encoder and decoder by matrix multiplication of the cross-attention map $A_C$ and its transpose $A_C^\top$.
By applying the objectives of $\mathcal{L}_{\text{feedback}}^E$ and $\mathcal{L}_{\text{feedback}}^D$ to minimize the gap between the guidance and self-attention maps, 
the temporal collapse of self-attention disappears.
This enables the model to precisely localize and classify action instances.
}
\label{fig:architecture}
\vspace{-10pt}
\end{figure*}

\section{Our Approach}
In this section, we elaborate our framework, Self-DETR, which resolves the temporal collapse problem of self-attention in DETR for TAD.
As shown in Fig.~\ref{fig:architecture}, Self-DETR follows the DETR~\cite{carion2020detr} architecture.
On top of it, our simple but powerful solution, self-feedback, works for guiding the self-attention layers with cross-attention map.
We first explain the original DETR and differences between the original and ours in the following subsection.
Afterwards, we introduce the motivation, specific method, and objective in sequence.

\subsection{Preliminary}
\noindent\textbf{DETR.} DETR~\cite{carion2020detr} is based on transformer~\cite{vaswani2017attention} architecture and thereby has two main components: encoder and decoder.
First, the transformer encoder of DETR is to learn global relations within input features.
DETR uses image features from CNN as the input tokens of the encoder.
There are multiple layers in the encoder to refine the input features gradually.
Each layer of the encoder consists of a self-attention module and multi-layer perceptron (MLP) with layer normalization and skip connection.

On the other hand, the decoder aims to learn the relationship between the encoder features and its own inputs, learnable embedding vectors.
They learn positional encoding for object detection so also called object queries. 
Similar to the encoder, it has several layers but also an additional cross-attention module in each layer to focus the relationship between the encoder features and the object queries.
In other words, the outputs of the encoder and object queries are fed into the decoder and reinforced by self-attention and cross-attention.
Finally, the output of the decoder is used to predict the class and the location of the object.

\vspace{3pt}
\noindent\textbf{Attention Mechanism.} Both the encoder and decoder have the attention modules~\cite{vaswani2017attention}.
They both need three variables so each has three linear layers to project inputs into three latent spaces.
The projected ones are called query $Q$, key $K$, and value $V$, respectively.
Then, the attention scores are calculated by matrix multiplication of $Q$ and the transpose of $K$ followed by the softmax activation function, which means how $Q$ and $K$ are similar.
By pooling $V$ with the scores followed by a linear projection, we can get the output of the attention modules.

Formally, $Q$, $K$, $V$ are in $\mathbb{R}^{N_Q \times D}$, $\mathbb{R}^{N_K \times D}$, $\mathbb{R}^{N_V \times D}$, respectively, where $N_Q$, $N_K$, $N_V$ are the lengths of $Q$, $K$, $V$, and $D$ is the number of channels.
Here, we assume that the number of channels for $Q$, $K$, $V$ are the same.
We can formulate the attention mechanism as follows:
\begin{equation}
    \begin{aligned}
    \text{Attention}(Q, K, V)=AV, \\
    A=\text{softmax}(\frac{QK^{\top}}{\sqrt{D}}),
    \label{eq:attention}
    \end{aligned}
\end{equation}
where A is the attention map, $A^{\top}$ indicates the transpose of $A$.
For the self-attention module, $Q$, $K$, $V$ are from the same input features.
On the other hand, $Q$ is from the decoder query embeddings, and $K$, $V$ from the encoder features in the cross-attention module.

\vspace{3pt}
\noindent\textbf{DETR for TAD.}
As for input to the model, we deploy features of a 3D CNN pre-trained on Kinetics~\cite{kay2017kinetics}.
Note that 3D CNN is fixed while training our model.
To extract the video features, each video is fed into the 3D CNN followed by global-average pooling in spatial dimensions so that only the temporal dimension remains.

Self-DETR follows the architecture of DAB-DETR~\cite{shilong2022dab_detr}, but decoder queries stand for action instances, called action queries.
Therefore, the decoder receives the refined video features from the encoder and relates them with action queries.
Finally, the output of the decoder passes through classification and regression heads, then final detection results are produced.

\begin{figure*}[t]
\centering
\includegraphics[width=17.40cm]{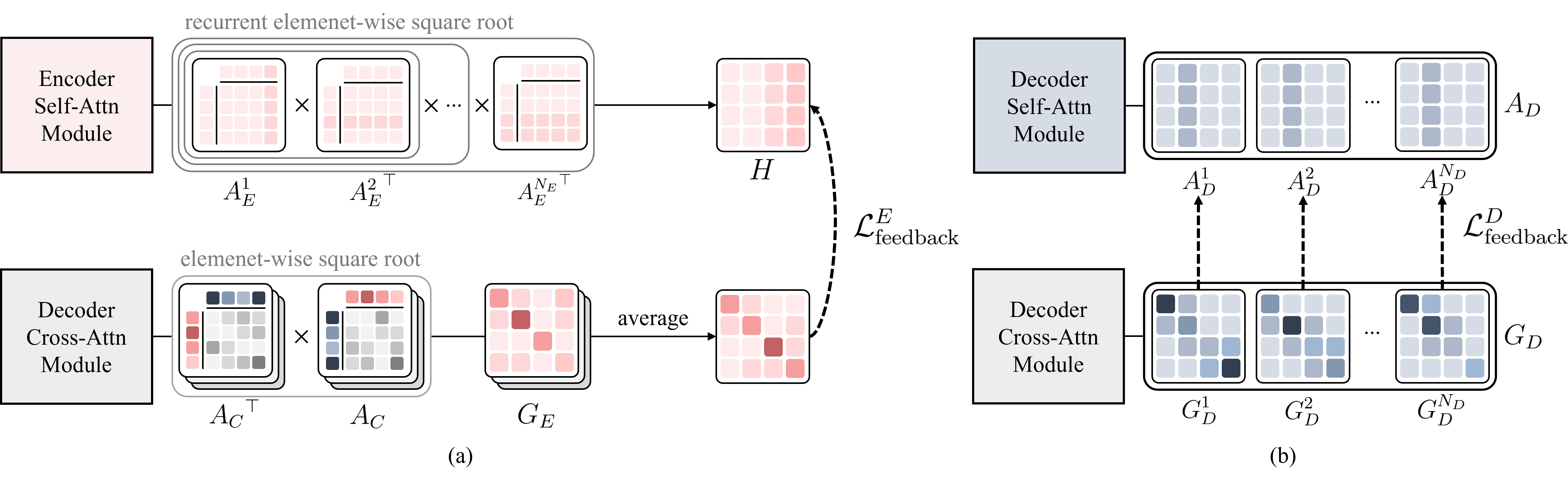}
\vspace{-25pt}
\caption{\textbf{Self-feedback methods.}
(a) For aggregating self-attention maps $A_E^i$ from multiple layers of the encoder, we use a series of matrix multiplication with recurrent element-wise square-root normalization.
Also, we average the guidance maps $G_E$ for the encoder from the cross-attention maps of the decoder.
The guidance map is applied on the aggregated self-attention map of the encoder $H$.
(b) On the other hand, the self-feedback for the decoder is provided layer by layer between the self-attention map $A_D^l$ and guidance map $G_D^l$.
}
\label{fig:feedback}
\vspace{-10pt}
\end{figure*}

\subsection{Guidance Maps}
\noindent\textbf{Motivation.} 
From~\cite{dong2021rank_collapse}, the pure attention module itself has the bias towards a rank-1 matrix exponentially with respect to the depth of the model without skip connections.
Unfortunately, the collapse problem also can be found in DETR for TAD as aforementioned Fig~\ref{fig:introduction} though it has skip connections.
This phenomenon implies that skip connections are not enough to slow down the collapse in TAD.
It is critical to DETR-based models because the self-attention modules are just skipped for the task without learning expressive relation.
Nonetheless, we observe that the cross-attention of DETR does not suffer from the temporal collapse by direct optimization from the objective.
The cross-attention map contains the entire relations between encoder features and decoder queries, so we process the map to get useful information for guiding self-attention.

In this paper, we emphasize to retain the standard attention in self-attention instead of replacing it.
The main benefit of keeping standard attention mechanism is that it introduces no inductive bias~\cite{dosovitskiy2021vit}.
Convolutions or deformable attention~\cite{xizhou2021deformable_detr} give a bias of locality so that it could lead the model to learn shortcuts.
In addition, it is eventually based on heuristics where pixels resemble their neighborhood.
The effort to remove biases sets the model free from limitations in learning~\cite{dosovitskiy2021vit}, which is aligned with the principal of DETR~\cite{carion2020detr}.

\vspace{3pt}
\noindent\textbf{Cross-Attention.}
The original goal of the cross-attention is to represent encoder features with the score of similarity between from decoder queries and encoder features.
Accordingly, we usually see the map as cross-relation between decoder queries $Q$ and encoder features $K$.
In addition, there are two perspectives to view the cross-attention map.
First, at the side of $Q$, they focus on the similar parts of $K$ when they are analogous.
On the other side of $K$, they are attended by the similar parts of $Q$ if they are related.

We emphasize the one's side by multiplying the cross-attention map and the transpose of it. 
When we denote the cross-attention map as $A_{C}$, we make the guidance map $G_D$ for the self-attention modules of the decoder as follows:
\begin{eqnarray}
G_D = \text{sqrt}(A_{C} {A_{C}}^{\top}),
\label{eq:QQ_guidance}
\end{eqnarray}
where $\text{sqrt}(\cdot)$ is element-wise square-root operation, $A_{C} \in \mathbb{R}^{L_q \times L_e}$ and $G_{D} \in \mathbb{R}^{L_q \times L_q}$ where $L_q$ and $L_e$ are the number of the query and encoder features.

Equivalently, we can obtain the guidance map $G_E$ for the self-attention modules of the encoder as follows:
\begin{eqnarray}
G_E = \text{sqrt}({A_{C}}^{\top} A_{C}),
\label{eq:KK_guidance}
\end{eqnarray}
where $G_E \in \mathbb{R}^{L_e \times L_e}$.

\subsection{Self-Feedback}
\label{subsec:self-feedback}
Now, we have the guidance maps $G_E$ and $G_D$ for the self-feedback to the self-attention modules of the encoder and decoder, respectively by Eq.~\ref{eq:QQ_guidance} and \ref{eq:KK_guidance}.
Still, there are more options to consider for how to handle the cross-attention and self-attention maps from the multiple layers of the encoder and decoder.
This consideration is quite important because the attention map on each layer exhibits different patterns as it goes deeper.
If we apply one single guidance map to force all self-attention maps to equally follow it, it will narrow down the diversity of representation.
From this perspective, we deliver how to choose or aggregate the cross-attention and self-attention maps before giving the feedback, as illustrated in Fig.~\ref{fig:feedback}.

\vspace{3pt}
\noindent\textbf{Encoder.}
Firstly, as for the encoder, we need to aggregate the self-attention maps of the encoder, on which we give the self-feedback.
There are two possible options for aggregation: 1) average pooling, 2) matrix multiplication.
One of the simplest ways is to average the self-attention maps from the encoder layers.
On the other hand, we also use matrix multiplication to aggregate them.
The stacked encoder layers can be viewed as the matrix multiplication of the subsequent self-attention maps when we assume there are no MLPs and skip connections.
In this sense, we aggregate the self-attention maps from the multiple layers of the encoder by a series of matrix multiplication.

Formally, let us denote the self-attention map in the $i$-th layer of the encoder as $A_E^i$.
Then we can describe the aggregated map as follows:
\begin{equation}
    \begin{aligned}
    H^{i} = \text{sqrt}(H^{i - 1} {A_E^{i}}^{\top}), 
    \label{eq:E_aggregation}
    \end{aligned}
\end{equation}
where $i = [2, 3, 4, ..., N_E]$, $H^1=A_E^1$. The final aggregated map $H=H^{N_E}$ with the number of encoder layers $N_E$.

As for the guidance map for the encoder, we simply average the cross-attention maps from the layers of the decoder.
Afterwards, we obtain the guidance map $G_E$ by Eq.~\ref{eq:QQ_guidance}.
With the given aggregated self-attention map $H$ of the encoder, and guidance map $G_E$ of the decoder,
we formulate the objective function of the self-feedback for the self-attention modules of the encoder as follows:
\begin{equation}
    \begin{aligned}
    \mathcal{L}_{\text{Feedback}}^{E} = D_{KL}(H~||~G_E),
    \label{eq:E_self}
    \end{aligned}
\end{equation}
where $D_{KL}$ is the Kullback–Leibler (KL) divergence.

\vspace{3pt}
\noindent\textbf{Decoder.}
The decoder also has multiple layers, and each layer has the self-attention and cross-attention modules.
On the same layer, the self-attention and cross-attention modules share the representations so their representation level is same.
Therefore, we do not aggregate the attention maps from the multiple layers.
Instead, we make the guidance map $G_D$ at each layer and give it to the self-attention module on the corresponding layer.

Formally, we define the self-attention map in the $l$-th layer of the decoder as $A^l_D$.
Similarly, let us denote the $G_D$ from the $l$-th layer of the cross-attention module as $G_D^l$.
We then formulate the cost function of the self-feedback for the self-attention modules of the decoder as follows:
\begin{equation}
    \begin{aligned}
    \mathcal{L}_{\text{Feedback}}^{D} = \sum_{l=1}^{N_D} D_{KL}(A^l_D~||~G_D^l),
    \label{eq:D_self}
    \end{aligned}
\end{equation}
where $N_D$ is the number of decoder layers.

\subsection{Objectives}
\noindent\textbf{DETR.}
Let us denote the ground-truths, and the $M$ predictions as $y$, $\hat{y}={\hat{y_i}_{i=1}^{M}}$, respectively.
For the bipartite matching between the ground-truth and prediction sets,
we define the optimal matching with the minimal cost to search for the permutation of $M$ elements $j \in J_M$  as follows:
\begin{equation}
    \begin{aligned}
    \hat{j} = \argmin_{j \in J_M} \sum_{i}^{M}\mathcal{L}_{\text{match}}(y_i, \hat{y}_{j(i)}),
    \label{eq:bipartite_matching}
    \end{aligned}
\end{equation}
where $L_{\text{match}(y_i, \hat{y}_{j(i)})}$ is a pair-wise matching cost between $y_i$ and the prediction with the index from $j(i)$, which outputs the index $i$ from the permutation $j$.

Next, let us denote each ground-truth action as $y_i = (c_i, t_i)$, where $c_i$ is the target class label with the background one $\emptyset$, and $t_i$ is the time intervals of the start and end times.
For the prediction with the index $j_{(i)}$, we define the probability of the class $c_i$ as $\hat{p}_{j(i)}(c_i)$ and the predicted time intervals as $\hat{t}_{\hat{j}(i)}$.
Then, we define $\mathcal{L}_{\text{match}}(y_i, \hat{y}_{j(i)})$ as below:
\begin{equation*}
    \begin{aligned}
    \mathcal{L}_{\text{match}}(y_i, \hat{y}_{j(i)}) = -\mathbbm{1}_{c_i \neq \emptyset}~\hat{p}_{j(i)}(c_i) + \mathbbm{1}_{c_i \neq \emptyset}~\mathcal{L}_{\text{reg}}(t_i, \hat{t}_{j(i)}),
    \label{eq:matching_cost}
    \end{aligned}
\end{equation*}
where $\mathcal{L}_{\text{reg}}(t_i, \hat{t}_{j(i)})$ is the regression loss between the ground-truth $t_i$ and the prediction $\hat{t}$ with the index $j(i)$.
The regression loss $\mathcal{L}_{\text{reg}}$ consists of L1 and Interaction-over-Union (IoU) losses as in the DETR-based methods~\cite{tan2021relaxed, liu2021tadtr, shi2022react}.
Finally, we formulate the main objective as following:
\begingroup
\small{
\begin{equation}
    \begin{aligned}
    \mathcal{L}_{\text{DETR}}(y, \hat{y}) = \sum_{i=1}^{M}[-\log{\hat{p}_{\hat{j}(i)}(c_i)} + \mathbbm{1}_{c_i \neq \emptyset}\mathcal{L}_{\text{reg}}(t_i, \hat{t}_{\hat{j}(i)})],
    \label{eq:detr_loss}
    \end{aligned}
\end{equation}
}
\endgroup
where $\hat{j}$ is the optimal assignment from Eq.~\ref{eq:bipartite_matching}.

\vspace{3pt}
\noindent\textbf{Full Objectives.}
To summarize the objectives for our framework, Self-DETR,
the full objective is can be described as below:
\begin{equation}
    \begin{aligned}
    \mathcal{L} = \mathcal{L}_{\text{DETR}} + \lambda _{E} \mathcal{L}_{\text{Feedback}}^{E} + \lambda _{D} \mathcal{L}_{\text{Feedback}}^{D},
    \label{eq:full_objective}
    \end{aligned}
\end{equation}
where $\lambda_{E}$ and $\lambda_{D}$ are the weights for the self-feedback losses for the encoder and decoder.
\begingroup
\setlength{\tabcolsep}{8.0pt} 
\renewcommand{\arraystretch}{1.0} 
\begin{table*}[t]
\centering
\begin{tabular}{l||ccccc|c||ccc|c}
    \hline\hline
    \multirow{2}{*}{Method} &
    \multicolumn{6}{c||}{THUMOS14} & \multicolumn{4}{c}{ActivityNet-v1.3} \\
    \cline{2-11}
    & $0.3$ & $0.4$ & $0.5$ & $0.6$ & $0.7$ & Avg. & $0.5$ & $0.75$ & $0.95$ & Avg.  \\
    \hline\hline 
    \rowcolor{gray!25}\multicolumn{11}{l}{\textit{\textbf{Standard Methods}}} \\
    \hline\hline
    BSN~\cite{lin2018bsn} & $53.5$ & $45.0$ & $36.9$ & $28.4$ & $20.0$ & $36.8$ & $46.46$ & $29.96$ & $8.02$ & $29.17$\\
    BMN~\cite{lin2019bmn} & $56.0$ & $47.4$ & $38.8$ & $29.7$ & $20.5$ & $38.5$ & $50.07$ & $34.78$ & $8.29$ & $33.85$ \\
    GTAD~\cite{xu2020gtad} & $54.5$ & $47.6$ & $40.2$ & $30.8$ & $23.4$ & $39.3$ & $50.36$ & $34.60$ & $9.02$ & $34.09$ \\
    BC-GNN~\cite{bai2020boundary} & $57.1$ & $49.1$ & $40.4$ & $31.2$ & $23.1$ & $40.2$ & $50.56$ & $34.75$ & $9.37$ & $34.26$ \\
    BSN++~\cite{su2021bsn++} & $59.9$ & $49.5$ & $41.3$ & $31.9$ & $22.8$ & $41.1$ & $51.27$ & $35.70$ & $8.33$ & $34.88$ \\
    IC \& IC~\cite{zhao2020bottom} & $53.9$ & $50.7$ & $45.4$ & $38.0$ & $28.5$ & $43.3$ & $43.47$ & $33.91$ & $9.21$ & $30.12$ \\
    MUSES~\cite{liu2021muse} & $\underline{68.9}$ & $\underline{64.0}$ & $\underline{56.9}$ & $\boldsymbol{46.3}$ & $31.0$ & $\underline{53.4}$ & $50.02$ & $34.97$ & $6.57$ & $33.99$ \\
    CSA~\cite{sridhar2021class} & $64.4$ & $58.0$ & $49.2$ & $38.2$ & $27.8$ & $47.5$ & $52.44$ & $\underline{36.69}$ & $5.18$ & $35.43$ \\
    PBRNet~\cite{liu2020progressive} & $58.5$ & $54.6$ & $51.3$ & $41.8$ & $29.5$ & $47.1$ & $53.96$ & $34.97$ & $8.98$ & $35.01$ \\
    VSGN~\cite{zhao2021stitching} & $66.7$ & $60.4$ & $52.4$ & $41.0$ & $30.4$ & $50.2$ & $52.38$ & $36.01$ & $8.37$ & $35.07$ \\
    ContextLoc~\cite{zhu2021contextloc} & $68.3$ & $63.8$ & $54.3$ & $41.8$ & $26.2$ & $50.9$ & $56.01$ & $35.19$ & $3.55$ & $34.23$ \\
    AFSD~\cite{lin2021salient} & $67.3$ & $62.4$ & $55.5$ & $43.7$ & $31.1$ & $52.0$ & $52.40$ & $35.30$ & $6.50$ & $34.40$ \\
    DCAN ~\cite{chen2022dcan} & $68.2$ & $62.7$ & $54.1$ & $43.9$ & $\boldsymbol{32.6}$ & $52.3$ & $51.78$ & $35.98$ & $\underline{9.45}$ & $35.39$ \\
    Zhu \textit{et al.}~\cite{zhu2022disentangled} & $\boldsymbol{72.1}$ & $\boldsymbol{65.9}$ & $\boldsymbol{57.0}$ & $\underline{44.2}$ & $28.5$ & $\boldsymbol{53.5}$ & $\boldsymbol{58.14}$ & $36.30$ & $6.16$ & $35.24$ \\
    RCL~\cite{wang2022rcl} & $70.1$ & $62.3$ & $52.9$ & $42.7$ & $30.7$ & $51.0$ & $54.19$ & $36.19$ & $9.17$ & $\underline{35.98}$ \\
    TAGS~\cite{nag2022tags} & $68.6$ & $63.8$ & $\boldsymbol{57.0}$ & $\boldsymbol{46.3}$ & $\underline{31.8}$ & $52.8$ & $\underline{56.30}$ & $\boldsymbol{36.80}$ & $\boldsymbol{9.60}$ & $\boldsymbol{36.50}$ \\
    \hline\hline
    \rowcolor{gray!25}\multicolumn{11}{l}{\textit{\textbf{DETR-based Methods}}} \\
    \hline\hline
    RTD-Net~\cite{tan2021relaxed} & $68.3$ & $62.3$ & $51.9$ & $38.8$ & $23.7$ & $49.0$ & $47.21$ & $30.68$ & $\boldsymbol{8.61}$ & $30.83$ \\
    TadTR~\cite{liu2021tadtr} & $62.4$ & $57.4$ & $49.2$ & $37.8$ & $26.3$ & $46.6$ & $49.10$ & $32.60$ & $8.50$ & $32.30$ \\
    ReAct~\cite{shi2022react} & $\underline{69.2}$ & $\underline{65.0}$ & $\underline{57.1}$ & $\boldsymbol{47.8}$ & $\boldsymbol{35.6}$ & $\underline{55.0}$ & $\underline{49.60}$ & $\underline{33.00}$ & $\underline{8.60}$ & $\underline{32.60}$ \\
    \hline
    Self-DETR & $\boldsymbol{74.6}$ & $\boldsymbol{69.5}$ & $\boldsymbol{60.0}$ & $\underline{47.6}$ & $\underline{31.8}$ & $\boldsymbol{56.7}$ & $\boldsymbol{52.25}$ & $\boldsymbol{33.67}$ & $8.40$ & $\boldsymbol{33.76}$ \\
    \hline\hline
\end{tabular}
\vspace{-5pt}
\caption{\textbf{The comparison results with the-state-of-the-art on THUMOS14 and ActivityNet-v1.3.} 
The table shows the evaluation results of the two types: standard and DETR-based models.
In THUMOS14, our model shows the state-of-the-art performance over all previous methods.
Also, our model outperforms the existing DETR-based methods on ActivityNet.
}
\label{tab:main}
\vspace{-10pt}
\end{table*}
\endgroup

\section{Experiments}
\subsection{Datasets}
Our experiments are conducted on the two challenging benchmarks of temporal action detection: THUMOS14~\cite{jiang2014thumos14} and ActivityNet-v1.3~\cite{caba2015activitynet}.

\textbf{THUMOS14} has 1,010 and 1,574 untrimmed videos as its validation and testing samples, respectively. Specifically, 200 and 212 videos have temporal annotations in the validation and testing sets, respectively. 
The dataset has 20 action classes. 
We use the validation set for training, and the testing one for evaluation.

\textbf{ActivityNet-v1.3} contains 19,994 videos with 200 action classes. 10024, 4926, and 5044 videos are for training, validation, and testing, respectively.

\subsection{Implementation Details}

\noindent\textbf{Architecture.}
We use the features of I3D~\cite{carreira2017i3d} pre-trained on Kinetics~\cite{kay2017kinetics}. 
In order for fair comparison, we are based on the size-modulated cross-attention module~\cite{shilong2022dab_detr} as the baselines~\cite{liu2021tadtr, shi2022react} deploy deformable attention~\cite{xizhou2021deformable_detr} which also uses the size-modulated attention.
Also, Self-DETR deploys learnable anchors and the way of updating predictions iteratively as done in ~\cite{liu2021tadtr, shi2022react}.
The number of layers of the encoder and decoder is $2$, and $4$, respectively.
The number of the queries is $40$.
We set the weights $\lambda_{E}$, $\lambda_{D}$ of the losses of the self-feedback for the encoder and decoder as $5$.

\vspace{3pt}
\noindent\textbf{Training.}
As for both datasets, we use Adam~\cite{kingma2014adam} as the optimizer with the batch size of 16.
For the input, we use 128 and 192 length of temporal features for THUMOS14 and ActivityNet-v1.3, respectively.

In THUMOS14, we train the framework for 120 epochs. 
The learning rate is decayed by $1/10$ when it reaches 80 and 100 epochs.
As for ActivityNet-v1.3, 20 epochs are taken for training. 
The learning rate decreases by cosine anealing with a warm-up of 5 epochs.
In addition, we resize the features of a video with linear interpolation to 192 length.

\vspace{3pt}
\noindent\textbf{Inference.} 
We slice the temporal features with a 128-length window with overlap of 32 for THUMOS14. 
As for ActivityNet-v1.3, we resize the features to 192 length as done in training.
Also, we use the top 100, and 200 predictions after non-maximum suppression (NMS) for the final localization results for ActivityNet-v1.3, and THUMOS14, respectively.
We use SoftNMS with the NMS threshoold of $0.40$.
For the class label, we fuse our classification scores with the top-1 video-level predictions of \cite{zhao2017cuhk} as done in ~\cite{tan2021relaxed, liu2021tadtr, shi2022react} for ActivityNet-v1.3. 

\subsection{Comparison with the State-of-the-Art}
We compare the-state-of-the-art methods to evaluate our framework, Self-DETR, on THUMOS14 and ActivityNet-v1.3 datasets.
Table.~\ref{tab:main} shows the comparison results with the state-of-the-art methods on THUMOS14 and ActivityNet-v1.3.
On THUMOS14, Self-DETR shows a new state-of-the-art performance over all of the existing approaches.
Compared to the standard methods which are not based on the set prediction, our model shows superior performances at all the IoU thresholds.
As for the DETR-based methods~\cite{tan2021relaxed, liu2021tadtr, shi2022react}, Self-DETR outperforms all the previous DETR-based methods in terms of the average mAP as well as the IoU thresholds of $0.3$, $0.4$, and $0.5$.

As for ActivityNet-v1.3, Self-DETR also shows a new state-of-the-art performance among the DETR-based approaches.
Interestingly, the APs at the $0.50$ and $0.95$ have almost reached the state-of-the-art of the standard methods.
This shows that the performance of the DETR-based methods has become comparable to the standard approaches.

\begin{figure}[t]
\centering
\includegraphics[width=8.35cm]{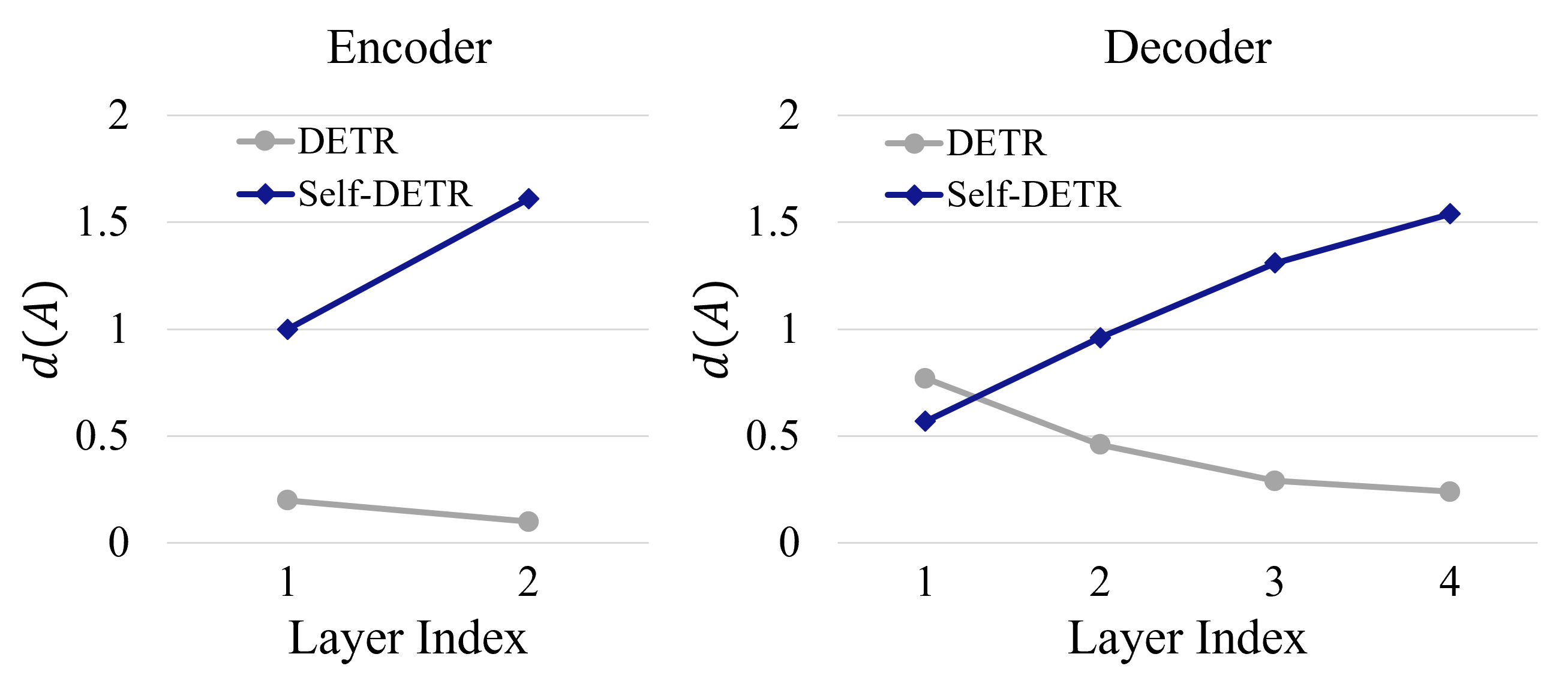}
\vspace{-20pt}
\caption{\textbf{Diversity of self-attention maps.}
In order to further analyze the effect of the self-feedback for the temporal collapse problem, we measure the diversity defined in Eq.~\ref{eq:diversity} of the self-attention maps.
}
\label{fig:diversity}
\vspace{-12pt}
\end{figure}

\subsection{Ablation Studies}
\noindent\textbf{Diversity of Self-Attention.} 
To further analyze the effect of the self-feedback, we can measure the diversity of self-attention maps according to~\cite{dong2021rank_collapse}.
The diversity $d(A)$ for the attention map $A$ is the measure of the closeness between the attention map and a rank-1 matrix as defined as below:
\begingroup
\small{
\begin{equation}
    \begin{aligned}
    d(A) = \| A - \boldsymbol{1}a^\top \|, \text{where}~a = \argmin_{a^{\prime}}\| A - \boldsymbol{1}a^{\prime\top} \|,
    \label{eq:diversity}
    \end{aligned}
\end{equation}
}
\endgroup
where $\| \cdot \|$ denotes the $\ell_1$,$\ell_\infty$-composite matrix norm, $a$, $a'$ are column vectors of the attention map $A$, and $\boldsymbol{1}$ is an all-ones vector.
Note that the rank of $\boldsymbol{1}a^\top$ is 1, and therefore, a smaller value of $d(A)$ means $A$ is closer to a rank-1 matrix.

Fig.~\ref{fig:diversity} shows the diversity on each layer of the encoder and decoder for the baseline DETR and Self-DETR.
The diversity is measured on the test set on THUMOS14 averaged over 160 randomly selected samples.
As the model depth gets deeper, the diversity of the baseline decreases close to $0$.
However, the diversity of Self-DETR does not fall down and even increases since the self-feedback loss guides the self-attention maps.

\begingroup
\setlength{\tabcolsep}{2.80pt} 
\renewcommand{\arraystretch}{1.0} 
\begin{table}[t]
	\centering
	\begin{tabular}{c|c|ccccc|c}
		\hline\hline
		$\mathcal{L}_{\text{Feedback}}^E$ & $\mathcal{L}_{\text{Feedback}}^D$ & $0.3$ & $0.4$ & $0.5$ & $0.6$ & $0.7$ & Avg. \\ \hline\hline
		$\cdot$ & $\cdot$ & $70.5$ & $64.3$ & $53.9$ & $39.3$ & $23.8$ & $50.3$ \\
        \checkmark & $\cdot$ & $73.5$ & $67.9$ & $57.2$ & $43.2$ & $26.6$ & $53.7$ \\
        $\cdot$ & \checkmark & $73.4$ & $67.4$ & $58.5$ & $44.5$ & $28.7$ & $54.4$ \\
        \checkmark & \checkmark & $\boldsymbol{74.5}$ & $\boldsymbol{69.5}$ & $\boldsymbol{60.0}$ & $\boldsymbol{47.6}$ & $\boldsymbol{31.8}$ & $\boldsymbol{56.7}$ \\
		\hline\hline
	\end{tabular}
    \vspace{-10pt}
	\caption{\textbf{Ablation on self-feedback.}
	To validate each self-feedback loss for the encoder and decoder, we have conducted experiments for ablating them on THUMOS14.}
	\label{tab:ablation}
    \vspace{-10.45pt}
\end{table}
\endgroup

\begingroup
\setlength{\tabcolsep}{3.00pt} 
\renewcommand{\arraystretch}{1.0} 
\begin{table}[t]
	\centering
	\begin{tabular}{c|c|ccccc|c}
		\hline\hline
		Encoder & Decoder & $0.3$ & $0.4$ & $0.5$ & $0.6$ & $0.7$ & Avg. \\ \hline\hline
		last & layer & $73.0$ & $67.2$ & $57.8$ & $44.9$ & $28.8$ & $54.3$ \\
        average & layer & $74.6$ & $68.6$ & $59.2$ & $46.0$ & $\boldsymbol{32.0}$ & $56.1$ \\
        matmul & layer & $\boldsymbol{74.6}$ & $\boldsymbol{69.5}$ & $\boldsymbol{60.0}$ & $\boldsymbol{47.6}$ & $31.8$ & $\boldsymbol{56.7}$ \\
        matmul & last & $72.2$ & $66.6$ & $56.6$ & $42.5$ & $27.8$ & $53.1$ \\
        matmul & average & $70.4$ & $63.8$ & $54.0$ & $41.5$ & $26.0$ & $51.1$ \\
		\hline\hline
	\end{tabular}
    \vspace{-10pt}
	\caption{\textbf{Aggregation methods for guidance.} 
	The table shows the comparison results of the aggregation methods for self-attention maps of encoder and decoder.}
	\label{tab:aggregation}
    \vspace{-15pt}
\end{table}
\endgroup

\vspace{3pt}
\noindent\textbf{Ablation on Self-Feedback.} 
In order to validate contribution of each self-feedback for the self-attention module of the encoder and decoder, we have conducted experiments for ablating the self-feedback on the THUMOS14 dataset.

Tab.~\ref{tab:ablation} shows the ablation study on the self-feedback.
As reported, each self-feedback for the encoder and decoder significantly improves the overall performance at all the IoU thresholds.
Moreover, when we use the self-feedback for both the encoder and decoder, the performance gain becomes larger reaching the state-of-the-art performance on THUMOS14.
Through this ablation study, each of the self-feedback losses $\mathcal{L}_{\text{Feedback}}^{E}$ and $\mathcal{L}_{\text{Feedback}}^{D}$ is crucial to relieve the temporal collapse problem of self-attention in DETR.

\vspace{3pt}
\noindent\textbf{Aggregation for Guidance.} 
To search for the optimal aggregation method, we conducted the experiments on THUMOS14 in Tab.~\ref{tab:aggregation}.
`last' means providing self-feedback only for the last self-attention layer.
`average' indicates average pooling self-attention maps and `matmul' is matrix multiplication.
Also, `layer' means self-feedback to each layer with guidance from the corresponding layer of the decoder.

In the table, `average' and `matmul' aggregation methods show superior performances over the `last' method.
As for the decoder, `layer' approach outperforms `last' and `average' methods as representation levels of self- and cross-attention on the same layer are well compatible.

\begingroup
\setlength{\tabcolsep}{5.00pt} 
\renewcommand{\arraystretch}{1.0} 
\begin{table}[t]
	\centering
	\begin{tabular}{l|ccccc|c}
		\hline\hline
		Method & $0.3$ & $0.4$ & $0.5$ & $0.6$ & $0.7$ & Avg. \\ \hline\hline
		baseline & $70.5$ & $64.3$ & $53.9$ & $39.3$ & $23.8$ & $50.3$ \\
        \hline
        relative & $69.1$ & $63.4$ & $53.8$ & $40.5$ & $24.6$ & $50.3$ \\
        identity & $69.1$ & $62.0$ & $52.9$ & $40.6$ & $24.6$ & $49.9$ \\
        diversity & $56.5$ & $49.5$ & $39.2$ & $26.2$ & $14.1$ & $37.1$ \\
        \hline
        cross-attn. & $\boldsymbol{74.6}$ & $\boldsymbol{69.5}$ & $\boldsymbol{60.0}$ & $\boldsymbol{47.6}$ & $\boldsymbol{31.8}$ & $\boldsymbol{56.7}$ \\
		\hline\hline
	\end{tabular}
    \vspace{-10pt}
	\caption{\textbf{Alternatives for self-feedback.} 
	The table shows results of alternatives for self-feedback.}
	\label{tab:alternatives}
    \vspace{-10.15pt}
\end{table}
\endgroup

\begingroup
\setlength{\tabcolsep}{5.5pt} 
\renewcommand{\arraystretch}{1.0} 
\begin{table}[t]
	\centering
	\begin{tabular}{c|c|ccccc|c}
		\hline\hline
		$\lambda_E$ & $\lambda_{D}$ & $0.3$ & $0.4$ & $0.5$ & $0.6$ & $0.7$ & Avg. \\ \hline\hline
        $1$ & $5$ & $73.1$ & $67.2$ & $58.6$ & $45.1$ & $29.7$ & $54.7$ \\
        $3$ & $5$ & $73.8$ & $67.8$ & $58.5$ & $46.4$ & $30.6$ & $55.4$ \\
        $5$ & $5$ & $\boldsymbol{74.6}$ & $\boldsymbol{69.5}$ & $\boldsymbol{60.0}$ & $\boldsymbol{47.6}$ & $31.8$ & $\boldsymbol{56.7}$ \\
        $5$ & $3$ & $73.7$ & $67.9$ & $59.1$ & $47.5$ & $\boldsymbol{32.1}$ & $56.1$ \\
        $5$ & $1$ & $73.4$ & $67.1$ & $57.7$ & $44.3$ & $28.3$ & $54.2$ \\
		\hline\hline
	\end{tabular}
    \vspace{-10pt}
	\caption{\textbf{Loss weights for self-feedback.}
	The table shows results depending on weights for self-feedback loss.}
	\label{tab:loss_weights}
    \vspace{-15pt}
\end{table}
\endgroup

\vspace{3pt}
\noindent\textbf{Alternatives for self-feedback.} 
Table.~\ref{tab:alternatives} shows the results for alternatives to the cross-attention maps for self-feedback on THUMOS14.
`relative' stands for deploying relative attention~\cite{shaw2018relative_attention} to self-attention of the encoder.
Also, `identity' means forcing the model to follow the identity matrix as guidance map, and `diversity' indicates we give the objective to maximize the diversity defined by Eq.~\ref{eq:diversity}.
The table shows that all alternatives do not bring a performance gain.
Hence, self-feedback of our guidance map from cross-attention is not just for regularization for diversity of self-attention but for learning expressive self-relation.

\vspace{3pt}
\noindent\textbf{Loss Weights of Self-Feedback.} 
Tab.~\ref{tab:loss_weights} shows performances for the different weights for the self-feedback losses for the encoder and decoder in THUMOS14.
It shows that the weight of $5$ is suitable for the encoder and decoder.
\section{Conclusion}
In this paper, we have explicitly discovered the temporal collapse problem of self-attention in DETR for TAD.
To alleviate the collapse, we have proposed a new framework, Self-DETR, which re-purposes cross-attention between the encoder and decoder in DETR to produce guidance map as self-feedback for self-attention.
Our extensive experiments have demonstrated that Self-DETR has resolved the collapse of self-attention preserving diversity of relation while reaching the state-of-the-art performance.

\vspace{3pt}
\noindent\textbf{Acknowledgement.}
This work was supported in part by MSIT/IITP (No. 2022-0-00680, 2019-0-00421, 2020-0- 01821, 2021-0-02068), and MSIT\&KNPA/KIPoT (Police Lab 2.0, No. 210121M06).

{\small
\bibliographystyle{ieee_fullname}
\bibliography{bibliography}

\begin{thebibliography}{10}\itemsep=-1pt

\bibitem{alwassel2018detad}
Humam Alwassel, Fabian~Caba Heilbron, Victor Escorcia, and Bernard Ghanem.
\newblock Diagnosing error in temporal action detectors.
\newblock In {\em Proceedings of the European conference on computer vision
  (ECCV)}, pages 256--272, 2018.

\bibitem{bai2020boundary}
Yueran Bai, Yingying Wang, Yunhai Tong, Yang Yang, Qiyue Liu, and Junhui Liu.
\newblock Boundary content graph neural network for temporal action proposal
  generation.
\newblock In {\em European Conference on Computer Vision}, pages 121--137.
  Springer, 2020.

\bibitem{buch2017sst}
Shyamal Buch, Victor Escorcia, Chuanqi Shen, Bernard Ghanem, and Juan
  Carlos~Niebles.
\newblock Sst: Single-stream temporal action proposals.
\newblock In {\em Proceedings of the IEEE conference on Computer Vision and
  Pattern Recognition}, pages 2911--2920, 2017.

\bibitem{caba2016fast}
Fabian Caba~Heilbron, Juan Carlos~Niebles, and Bernard Ghanem.
\newblock Fast temporal activity proposals for efficient detection of human
  actions in untrimmed videos.
\newblock In {\em Proceedings of the IEEE conference on computer vision and
  pattern recognition}, pages 1914--1923, 2016.

\bibitem{carion2020detr}
Nicolas Carion, Francisco Massa, Gabriel Synnaeve, Nicolas Usunier, Alexander
  Kirillov, and Sergey Zagoruyko.
\newblock End-to-end object detection with transformers.
\newblock In {\em Computer Vision--ECCV 2020: 16th European Conference,
  Glasgow, UK, August 23--28, 2020, Proceedings, Part I 16}, pages 213--229.
  Springer, 2020.

\bibitem{carreira2017i3d}
Joao Carreira and Andrew Zisserman.
\newblock Quo vadis, action recognition? a new model and the kinetics dataset.
\newblock In {\em proceedings of the IEEE Conference on Computer Vision and
  Pattern Recognition}, pages 6299--6308, 2017.

\bibitem{chao2018tal-net}
Yu-Wei Chao, Sudheendra Vijayanarasimhan, Bryan Seybold, David~A Ross, Jia
  Deng, and Rahul Sukthankar.
\newblock Rethinking the faster r-cnn architecture for temporal action
  localization.
\newblock In {\em Proceedings of the IEEE Conference on Computer Vision and
  Pattern Recognition}, pages 1130--1139, 2018.

\bibitem{chen2022dcan}
Guo Chen, Yin-Dong Zheng, Limin Wang, and Tong Lu.
\newblock Dcan: improving temporal action detection via dual context
  aggregation.
\newblock In {\em Proceedings of the AAAI Conference on Artificial
  Intelligence}, volume~36, pages 248--257, 2022.

\bibitem{dai2017tcn}
Xiyang Dai, Bharat Singh, Guyue Zhang, Larry~S Davis, and Yan Qiu~Chen.
\newblock Temporal context network for activity localization in videos.
\newblock In {\em Proceedings of the IEEE International Conference on Computer
  Vision}, pages 5793--5802, 2017.

\bibitem{dong2021rank_collapse}
Yihe Dong, Jean-Baptiste Cordonnier, and Andreas Loukas.
\newblock Attention is not all you need: Pure attention loses rank doubly
  exponentially with depth.
\newblock In {\em International Conference on Machine Learning}, pages
  2793--2803. PMLR, 2021.

\bibitem{dosovitskiy2021vit}
Alexey Dosovitskiy, Lucas Beyer, Alexander Kolesnikov, Dirk Weissenborn,
  Xiaohua Zhai, Thomas Unterthiner, Mostafa Dehghani, Matthias Minderer, Georg
  Heigold, Sylvain Gelly, Jakob Uszkoreit, and Neil Houlsby.
\newblock An image is worth 16x16 words: Transformers for image recognition at
  scale.
\newblock In {\em 9th International Conference on Learning Representations,
  {ICLR} 2021, Virtual Event, Austria, May 3-7, 2021}. OpenReview.net, 2021.

\bibitem{escorcia2016dap}
Victor Escorcia, Fabian~Caba Heilbron, Juan~Carlos Niebles, and Bernard Ghanem.
\newblock Daps: Deep action proposals for action understanding.
\newblock In {\em European Conference on Computer Vision}, pages 768--784.
  Springer, 2016.

\bibitem{caba2015activitynet}
Bernard~Ghanem Fabian Caba~Heilbron, Victor~Escorcia and Juan~Carlos Niebles.
\newblock Activitynet: A large-scale video benchmark for human activity
  understanding.
\newblock In {\em Proceedings of the IEEE Conference on Computer Vision and
  Pattern Recognition}, pages 961--970, 2015.

\bibitem{gao2018ctap}
Jiyang Gao, Kan Chen, and Ram Nevatia.
\newblock Ctap: Complementary temporal action proposal generation.
\newblock In {\em Proceedings of the European Conference on Computer Vision
  (ECCV)}, pages 68--83, 2018.

\bibitem{gao2017turn}
Jiyang Gao, Zhenheng Yang, Kan Chen, Chen Sun, and Ram Nevatia.
\newblock Turn tap: Temporal unit regression network for temporal action
  proposals.
\newblock In {\em Proceedings of the IEEE International Conference on Computer
  Vision}, pages 3628--3636, 2017.

\bibitem{girshick2015fast-rcnn}
Ross Girshick.
\newblock Fast r-cnn.
\newblock In {\em Proceedings of the IEEE international conference on computer
  vision}, pages 1440--1448, 2015.

\bibitem{girshick2014rcnn}
Ross Girshick, Jeff Donahue, Trevor Darrell, and Jitendra Malik.
\newblock Rich feature hierarchies for accurate object detection and semantic
  segmentation.
\newblock In {\em Proceedings of the IEEE conference on computer vision and
  pattern recognition}, pages 580--587, 2014.

\bibitem{jiang2014thumos14}
Y.-G. Jiang, J. Liu, A. Roshan~Zamir, G. Toderici, I. Laptev, M. Shah, and R.
  Sukthankar.
\newblock {THUMOS} challenge: Action recognition with a large number of
  classes.
\newblock \url{http://crcv.ucf.edu/THUMOS14/}, 2014.

\bibitem{kang2023amnet}
Tae-Kyung Kang, Gun-Hee Lee, Kyung-Min Jin, and Seong-Whan Lee.
\newblock Action-aware masking network with group-based attention for temporal
  action localization.
\newblock In {\em Proceedings of the IEEE/CVF Winter Conference on Applications
  of Computer Vision}, pages 6058--6067, 2023.

\bibitem{kay2017kinetics}
Will Kay, Joao Carreira, Karen Simonyan, Brian Zhang, Chloe Hillier, Sudheendra
  Vijayanarasimhan, Fabio Viola, Tim Green, Trevor Back, Paul Natsev, et~al.
\newblock The kinetics human action video dataset.
\newblock {\em arXiv preprint arXiv:1705.06950}, 2017.

\bibitem{kim2019coarsefine}
Ji{-}Hwan Kim and Jae{-}Pil Heo.
\newblock Learning coarse and fine features for precise temporal action
  localization.
\newblock {\em {IEEE} Access}, 7:149797--149809, 2019.

\bibitem{kingma2014adam}
Diederik~P Kingma and Jimmy Ba.
\newblock Adam: A method for stochastic optimization.
\newblock {\em arXiv preprint arXiv:1412.6980}, 2014.

\bibitem{lei2021moment-detr}
Jie Lei, Tamara~L Berg, and Mohit Bansal.
\newblock Detecting moments and highlights in videos via natural language
  queries.
\newblock {\em Advances in Neural Information Processing Systems},
  34:11846--11858, 2021.

\bibitem{lin2021salient}
Chuming Lin, Chengming Xu, Donghao Luo, Yabiao Wang, Ying Tai, Chengjie Wang,
  Jilin Li, Feiyue Huang, and Yanwei Fu.
\newblock Learning salient boundary feature for anchor-free temporal action
  localization.
\newblock In {\em Proceedings of the IEEE/CVF Conference on Computer Vision and
  Pattern Recognition}, pages 3320--3329, 2021.

\bibitem{lin2019bmn}
Tianwei Lin, Xiao Liu, Xin Li, Errui Ding, and Shilei Wen.
\newblock Bmn: Boundary-matching network for temporal action proposal
  generation.
\newblock In {\em Proceedings of the IEEE/CVF International Conference on
  Computer Vision}, pages 3889--3898, 2019.

\bibitem{lin2018bsn}
Tianwei Lin, Xu Zhao, Haisheng Su, Chongjing Wang, and Ming Yang.
\newblock Bsn: Boundary sensitive network for temporal action proposal
  generation.
\newblock In {\em Proceedings of the European Conference on Computer Vision
  (ECCV)}, pages 3--19, 2018.

\bibitem{liu2020progressive}
Qinying Liu and Zilei Wang.
\newblock Progressive boundary refinement network for temporal action
  detection.
\newblock In {\em Proceedings of the AAAI Conference on Artificial
  Intelligence}, volume~34, pages 11612--11619, 2020.

\bibitem{shilong2022dab_detr}
Shilong Liu, Feng Li, Hao Zhang, Xiao Yang, Xianbiao Qi, Hang Su, Jun Zhu, and
  Lei Zhang.
\newblock {DAB-DETR:} dynamic anchor boxes are better queries for {DETR}.
\newblock In {\em The Tenth International Conference on Learning
  Representations, {ICLR} 2022, Virtual Event, April 25-29, 2022}.
  OpenReview.net, 2022.

\bibitem{liu2022empirical}
Xiaolong Liu, Song Bai, and Xiang Bai.
\newblock An empirical study of end-to-end temporal action detection.
\newblock In {\em Proceedings of the IEEE/CVF Conference on Computer Vision and
  Pattern Recognition}, pages 20010--20019, 2022.

\bibitem{liu2021muse}
Xiaolong Liu, Yao Hu, Song Bai, Fei Ding, Xiang Bai, and Philip~HS Torr.
\newblock Multi-shot temporal event localization: a benchmark.
\newblock In {\em Proceedings of the IEEE/CVF Conference on Computer Vision and
  Pattern Recognition}, pages 12596--12606, 2021.

\bibitem{liu2021tadtr}
Xiaolong Liu, Qimeng Wang, Yao Hu, Xu Tang, Shiwei Zhang, Song Bai, and Xiang
  Bai.
\newblock End-to-end temporal action detection with transformer.
\newblock {\em arXiv preprint arXiv:2106.10271}, 2021.

\bibitem{meng2021conditional_detr}
Depu Meng, Xiaokang Chen, Zejia Fan, Gang Zeng, Houqiang Li, Yuhui Yuan, Lei
  Sun, and Jingdong Wang.
\newblock Conditional detr for fast training convergence.
\newblock In {\em Proceedings of the IEEE/CVF International Conference on
  Computer Vision}, pages 3651--3660, 2021.

\bibitem{moon2023qd-detr}
WonJun Moon, Sangeek Hyun, SangUk Park, Dongchan Park, and Jae-Pil Heo.
\newblock Query-dependent video representation for moment retrieval and
  highlight detection.
\newblock In {\em Proceedings of the IEEE/CVF Conference on Computer Vision and
  Pattern Recognition}, pages 23023--23033, 2023.

\bibitem{nag2022tags}
Sauradip Nag, Xiatian Zhu, Yi-Zhe Song, and Tao Xiang.
\newblock Proposal-free temporal action detection via global segmentation mask
  learning.
\newblock In {\em Computer Vision--ECCV 2022: 17th European Conference, Tel
  Aviv, Israel, October 23--27, 2022, Proceedings, Part III}, pages 645--662.
  Springer, 2022.

\bibitem{qing2021temporal}
Zhiwu Qing, Haisheng Su, Weihao Gan, Dongliang Wang, Wei Wu, Xiang Wang, Yu
  Qiao, Junjie Yan, Changxin Gao, and Nong Sang.
\newblock Temporal context aggregation network for temporal action proposal
  refinement.
\newblock In {\em Proceedings of the IEEE/CVF conference on computer vision and
  pattern recognition}, pages 485--494, 2021.

\bibitem{qiu2018etp}
Haonan Qiu, Yingbin Zheng, Hao Ye, Yao Lu, Feng Wang, and Liang He.
\newblock Precise temporal action localization by evolving temporal proposals.
\newblock In {\em Proceedings of the 2018 ACM on International Conference on
  Multimedia Retrieval}, pages 388--396. ACM, 2018.

\bibitem{ren2015faster-rcnn}
Shaoqing Ren, Kaiming He, Ross Girshick, and Jian Sun.
\newblock Faster r-cnn: Towards real-time object detection with region proposal
  networks.
\newblock In {\em Advances in neural information processing systems}, pages
  91--99, 2015.

\bibitem{byungseok2022sparse_detr}
Byungseok Roh, Jaewoong Shin, Wuhyun Shin, and Saehoon Kim.
\newblock Sparse {DETR:} efficient end-to-end object detection with learnable
  sparsity.
\newblock In {\em The Tenth International Conference on Learning
  Representations, {ICLR} 2022, Virtual Event, April 25-29, 2022}.
  OpenReview.net, 2022.

\bibitem{shaw2018relative_attention}
Peter Shaw, Jakob Uszkoreit, and Ashish Vaswani.
\newblock Self-attention with relative position representations.
\newblock In Marilyn~A. Walker, Heng Ji, and Amanda Stent, editors, {\em
  Proceedings of the 2018 Conference of the North American Chapter of the
  Association for Computational Linguistics: Human Language Technologies,
  NAACL-HLT, New Orleans, Louisiana, USA, June 1-6, 2018, Volume 2 (Short
  Papers)}, pages 464--468. Association for Computational Linguistics, 2018.

\bibitem{shi2022react}
Dingfeng Shi, Yujie Zhong, Qiong Cao, Jing Zhang, Lin Ma, Jia Li, and Dacheng
  Tao.
\newblock React: Temporal action detection with relational queries.
\newblock In {\em Computer Vision--ECCV 2022: 17th European Conference, Tel
  Aviv, Israel, October 23--27, 2022, Proceedings, Part X}, pages 105--121.
  Springer, 2022.

\bibitem{shou2017cdc}
Zheng Shou, Jonathan Chan, Alireza Zareian, Kazuyuki Miyazawa, and Shih-Fu
  Chang.
\newblock Cdc: Convolutional-de-convolutional networks for precise temporal
  action localization in untrimmed videos.
\newblock In {\em Proceedings of the IEEE Conference on Computer Vision and
  Pattern Recognition}, pages 5734--5743, 2017.

\bibitem{shou2016scnn}
Zheng Shou, Dongang Wang, and Shih-Fu Chang.
\newblock Temporal action localization in untrimmed videos via multi-stage
  cnns.
\newblock In {\em Proceedings of the IEEE Conference on Computer Vision and
  Pattern Recognition}, pages 1049--1058, 2016.

\bibitem{sridhar2021class}
Deepak Sridhar, Niamul Quader, Srikanth Muralidharan, Yaoxin Li, Peng Dai, and
  Juwei Lu.
\newblock Class semantics-based attention for action detection.
\newblock In {\em Proceedings of the IEEE/CVF International Conference on
  Computer Vision}, pages 13739--13748, 2021.

\bibitem{su2021bsn++}
Haisheng Su, Weihao Gan, Wei Wu, Yu Qiao, and Junjie Yan.
\newblock Bsn++: Complementary boundary regressor with scale-balanced relation
  modeling for temporal action proposal generation.
\newblock In {\em Proceedings of the AAAI Conference on Artificial
  Intelligence}, volume~35, pages 2602--2610, 2021.

\bibitem{tan2021relaxed}
Jing Tan, Jiaqi Tang, Limin Wang, and Gangshan Wu.
\newblock Relaxed transformer decoders for direct action proposal generation.
\newblock In {\em Proceedings of the IEEE/CVF International Conference on
  Computer Vision (ICCV)}, pages 13526--13535, October 2021.

\bibitem{tran2015c3d}
Du Tran, Lubomir Bourdev, Rob Fergus, Lorenzo Torresani, and Manohar Paluri.
\newblock Learning spatiotemporal features with 3d convolutional networks.
\newblock In {\em Proceedings of the IEEE international conference on computer
  vision}, pages 4489--4497, 2015.

\bibitem{vaswani2017attention}
Ashish Vaswani, Noam Shazeer, Niki Parmar, Jakob Uszkoreit, Llion Jones,
  Aidan~N Gomez, {\L}ukasz Kaiser, and Illia Polosukhin.
\newblock Attention is all you need.
\newblock In {\em Advances in neural information processing systems}, pages
  5998--6008, 2017.

\bibitem{wang2022rcl}
Qiang Wang, Yanhao Zhang, Yun Zheng, and Pan Pan.
\newblock Rcl: Recurrent continuous localization for temporal action detection.
\newblock In {\em Proceedings of the IEEE/CVF Conference on Computer Vision and
  Pattern Recognition (CVPR)}, pages 13566--13575, June 2022.

\bibitem{xiong2017tag}
Yuanjun Xiong, Yue Zhao, Limin Wang, Dahua Lin, and Xiaoou Tang.
\newblock A pursuit of temporal accuracy in general activity detection.
\newblock {\em arXiv preprint arXiv:1703.02716}, 2017.

\bibitem{xu2017rc3d}
Huijuan Xu, Abir Das, and Kate Saenko.
\newblock R-c3d: Region convolutional 3d network for temporal activity
  detection.
\newblock In {\em Proceedings of the IEEE international conference on computer
  vision}, pages 5783--5792, 2017.

\bibitem{xu2020gtad}
Mengmeng Xu, Chen Zhao, David~S Rojas, Ali Thabet, and Bernard Ghanem.
\newblock G-tad: Sub-graph localization for temporal action detection.
\newblock In {\em Proceedings of the IEEE/CVF Conference on Computer Vision and
  Pattern Recognition}, pages 10156--10165, 2020.

\bibitem{yeung2016frame-glimpses}
Serena Yeung, Olga Russakovsky, Greg Mori, and Li Fei-Fei.
\newblock End-to-end learning of action detection from frame glimpses in
  videos.
\newblock In {\em Proceedings of the IEEE Conference on Computer Vision and
  Pattern Recognition}, pages 2678--2687, 2016.

\bibitem{yuan2016score-distribution}
Jun Yuan, Bingbing Ni, Xiaokang Yang, and Ashraf~A Kassim.
\newblock Temporal action localization with pyramid of score distribution
  features.
\newblock In {\em Proceedings of the IEEE Conference on Computer Vision and
  Pattern Recognition}, pages 3093--3102, 2016.

\bibitem{yuan2017maximul-sum}
Zehuan Yuan, Jonathan~C Stroud, Tong Lu, and Jia Deng.
\newblock Temporal action localization by structured maximal sums.
\newblock In {\em Proceedings of the IEEE Conference on Computer Vision and
  Pattern Recognition}, pages 3684--3692, 2017.

\bibitem{zeng2019pgcn}
Runhao Zeng, Wenbing Huang, Mingkui Tan, Yu Rong, Peilin Zhao, Junzhou Huang,
  and Chuang Gan.
\newblock Graph convolutional networks for temporal action localization.
\newblock In {\em Proceedings of the IEEE/CVF International Conference on
  Computer Vision}, pages 7094--7103, 2019.

\bibitem{zhang2022actionformer}
Chen-Lin Zhang, Jianxin Wu, and Yin Li.
\newblock Actionformer: Localizing moments of actions with transformers.
\newblock In {\em Computer Vision--ECCV 2022: 17th European Conference, Tel
  Aviv, Israel, October 23--27, 2022, Proceedings, Part IV}, pages 492--510.
  Springer, 2022.

\bibitem{zhang2022dino}
Hao Zhang, Feng Li, Shilong Liu, Lei Zhang, Hang Su, Jun Zhu, Lionel Ni, and
  Harry Shum.
\newblock Dino: Detr with improved denoising anchor boxes for end-to-end object
  detection.
\newblock In {\em International Conference on Learning Representations}, 2022.

\bibitem{zhao2021stitching}
Chen Zhao, Ali~K Thabet, and Bernard Ghanem.
\newblock Video self-stitching graph network for temporal action localization.
\newblock In {\em Proceedings of the IEEE/CVF International Conference on
  Computer Vision}, pages 13658--13667, 2021.

\bibitem{zhao2020bottom}
Peisen Zhao, Lingxi Xie, Chen Ju, Ya Zhang, Yanfeng Wang, and Qi Tian.
\newblock Bottom-up temporal action localization with mutual regularization.
\newblock In {\em European Conference on Computer Vision}, pages 539--555.
  Springer, 2020.

\bibitem{zhao2017ssn}
Yue Zhao, Yuanjun Xiong, Limin Wang, Zhirong Wu, Xiaoou Tang, and Dahua Lin.
\newblock Temporal action detection with structured segment networks.
\newblock In {\em Proceedings of the IEEE International Conference on Computer
  Vision}, pages 2914--2923, 2017.

\bibitem{zhao2017cuhk}
Y Zhao, B Zhang, Z Wu, S Yang, L Zhou, S Yan, L Wang, Y Xiong, D Lin, Y Qiao,
  et~al.
\newblock Cuhk \& ethz \& siat submission to activitynet challenge 2017.
\newblock {\em arXiv preprint arXiv:1710.08011}, 8, 2017.

\bibitem{xizhou2021deformable_detr}
Xizhou Zhu, Weijie Su, Lewei Lu, Bin Li, Xiaogang Wang, and Jifeng Dai.
\newblock Deformable {DETR:} deformable transformers for end-to-end object
  detection.
\newblock In {\em 9th International Conference on Learning Representations,
  {ICLR} 2021, Virtual Event, Austria, May 3-7, 2021}. OpenReview.net, 2021.

\bibitem{zhu2021contextloc}
Zixin Zhu, Wei Tang, Le Wang, Nanning Zheng, and Gang Hua.
\newblock Enriching local and global contexts for temporal action localization.
\newblock In {\em Proceedings of the IEEE/CVF International Conference on
  Computer Vision (ICCV)}, pages 13516--13525, October 2021.

\bibitem{zhu2022disentangled}
Zixin Zhu, Le Wang, Wei Tang, Ziyi Liu, Nanning Zheng, and Gang Hua.
\newblock Learning disentangled classification and localization representations
  for temporal action localization.
\newblock In {\em Proceedings of the AAAI Conference on Artificial
  Intelligence}, volume~2, 2022.

\end{thebibliography}
}

\renewcommand{\thepage}{A\arabic{page}}  
\renewcommand{\thesection}{A}   
\renewcommand{\thetable}{A\arabic{table}}   
\renewcommand{\thefigure}{A\arabic{figure}}

\setcounter{section}{0}
\setcounter{table}{0}
\setcounter{figure}{0}

\newpage



\section{Appendix}
\subsection{Additional Details}
\noindent\textbf{More Training Details.}
As mentioned in the paper, $\mathcal{L}_{\text{reg}}$ of the paper in $\mathcal{L}_{\text{DETR}}$ defined in Eq.~8 of the paper consists of two types losses: L1 and Interaction-over-Union (IoU) losses.
We use the weights for L1 and IoU losses as $2$ and $5$, respectively as done in the baselines~\cite{liu2021tadtr, shi2022react} for both benchmarks.
Moreover, when we define $-\log \hat{p}_{\hat{j}(i)}(c_i)$ in $\mathcal{L}_{\text{DETR}}$ as $\mathcal{L}_{\text{cls}}$, we use the weight for $\mathcal{L}_{\text{cls}}$ as $2$ for both datasets.
The initial learning rates are $2 \times 10^-4$ and $1 \times 10^-4$ for THUMOS14 and ActivityNet.

\begin{figure}[t]
\centering
\includegraphics[width=8.35cm]{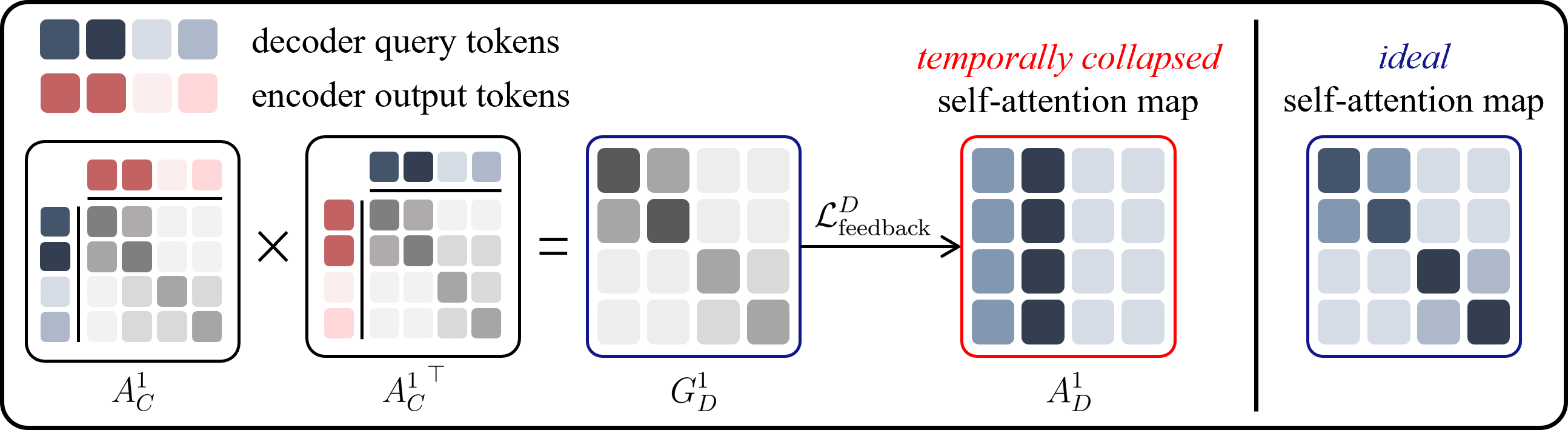}
\caption{\textbf{Further explanation for guidance Map of the decoder.} The figure illustrates an example of constructing guidance map for self-attention of the decoder.
}
\label{fig:rebuttal_guidance}
\end{figure}

\vspace{3pt}
\noindent\textbf{Further Explanation for Guidance Map.}
The guidance mechanisms are the same for both the encoder and the decoder.
Let us explain the guidance map specifically for the decoder's self-attention with an example in Fig.~\ref{fig:rebuttal_guidance}.
If the 1st and 2nd decoder queries are similar, 
they will attend similar encoder tokens, as the elements $a_{11}$, $a_{12}$, $a_{21}$ and $a_{22}$ with high values in the cross-attention map $A_{C}^1$ in the figure.
Then, the elements $g_{12}$ and $g_{21}$ in the guidance map $G_{D}^1$, calculated by matrix multiplication of $A_{C}^1$ and its transpose, will have high values, indicating high correlation b/w the 1st and 2nd queries.
Let us review the \textit{ideal} self-attention: the elements $a_{12}$ and $a_{21}$ in the self-attention map $A_{D}^1$ should also have high values if the 1st and 2nd decoder queries are similar.
It implies $G_{D}^1$ is analogous to \textit{ideal} case of $A_{D}^1$ so it can be a guidance to \textit{temporally collapsed} $A_{D}^1$.

\begin{figure}[t]
\centering
\includegraphics[width=8.35cm]{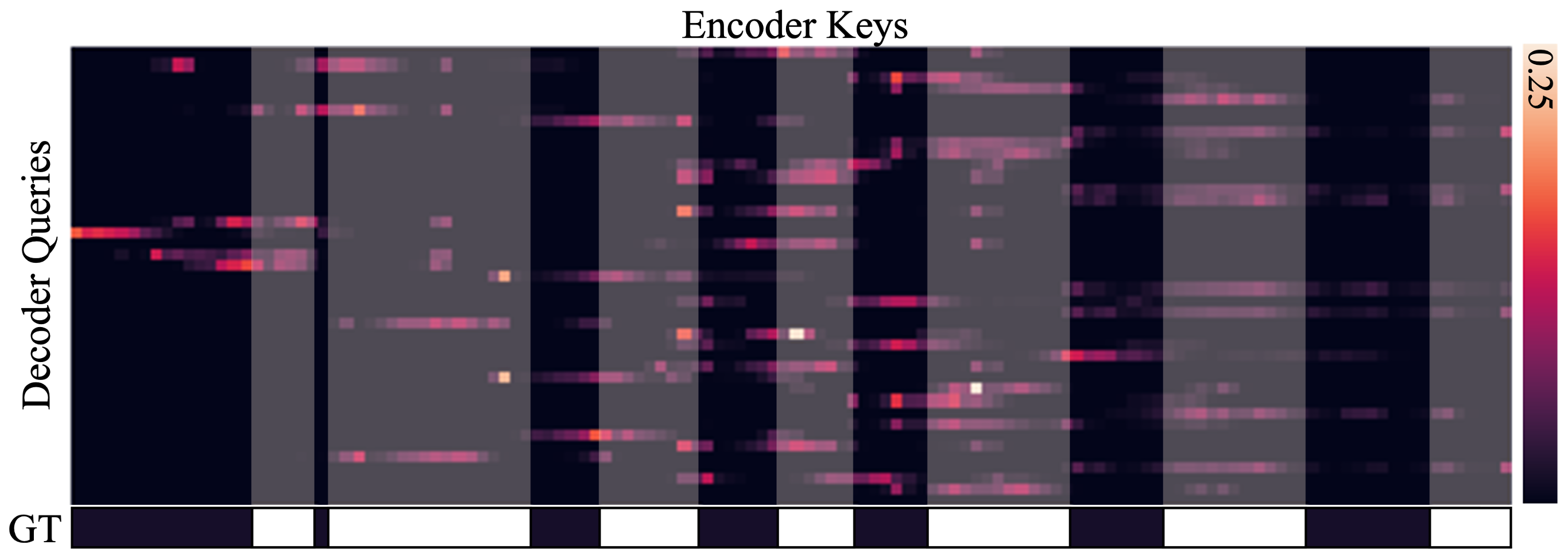}
\caption{\textbf{Visualization of the decoder cross-attention map.} The figure depicts a cross-attention map of the final decoder layer for a validation video in THUMOS14.
}
\label{fig:rebuttal_cross_attention}
\end{figure}

\vspace{3pt}
\noindent\textbf{Motivation behind the Design.}
Decoder queries do not always attend foreground features exclusively; they often include both foreground and background features simultaneously, as in Fig.~\ref{fig:rebuttal_cross_attention}.
Background regions serve two essential purposes for TAD.
Firstly, they define the boundaries of the actions through the surrounding background frames.
Secondly, background features provide contextual information about the actions.
Thus, a guidance map exhibiting a strong correlation between foreground and background encoder features is not only intuitive but also valuable.

From sound self-attention maps in DETR of object detection as in Fig.~1(a) of the paper, we observe two key characteristics: 1) correlation between adjacent tokens, 2) diversity.
First of all, based on the further explanation for guidance map in the previous paragraph, the cross-attention map encompasses correlations within encoder features or within decoder queries.
This motivates us to make references for self-attention maps using cross-relation.
Furthermore, we can introduce diversity to self-attention as our guidance maps ensure high diversity.
This is because the diversity of guidance map follows high diversity of the cross-attention map.
Mathematically, it is trivial that $\text{rank}A=\text{rank}AA^{\top}=\text{rank}A^{\top}A$ for a real matrix $A$.

\subsection{Additional Results}
\noindent\textbf{Number of Layers.}
Tab.~\ref{tab:number_of_layers} shows performances with various numbers of layers for the encoder and decoder.
Note that we use as default $2$ and $4$ layers for the encoder and decoder, respectively.
As seen in the table, the performance consistently decreases when using a small number of layers than the default setting.

\vspace{3pt}
\noindent\textbf{Ablation on Collapsed Self-Attention.} 
As mentioned in the paper, we argue that the collapsed self-attention modules in the encoder and decoder will play no role for the task.
Tab.~\ref{tab:ablation_on_collpased_self_attention} shows performances of ablation on the collapsed self-attention modules.
To ablate self-attention of the encoder, we remove the entire encoder.
As for the decoder, we just remove the self-attention modules.

As seen in the table, the performance drop is quite marginal when we ablate the entire encoder or decoder self-attention.
From this result, we find that the collapsed self-attention modules hardly help the model to solve TAD.

\begingroup
\setlength{\tabcolsep}{2.80pt} 
\renewcommand{\arraystretch}{1.0} 
\begin{table}[t]
	\centering
	\begin{tabular}{c|c|ccccc|c}
		\hline\hline
		Encoder & Decoder & $0.3$ & $0.4$ & $0.5$ & $0.6$ & $0.7$ & Avg. \\ \hline\hline
		$1$ & $4$ & $74.4$ & $69.2$ & $59.6$ & $45.8$ & $29.2$ & $55.6$ \\
        $2$ & $4$ & $\boldsymbol{74.6}$ & $\boldsymbol{69.5}$ & $\boldsymbol{60.0}$ & $\boldsymbol{47.6}$ & $\boldsymbol{31.8}$ & $\boldsymbol{56.7}$ \\
		$2$ & $3$ & $74.1$ & $67.4$ & $58.6$ & $45.3$ & $29.9$ & $55.1$ \\
		$2$ & $2$ & $67.5$ & $61.9$ & $51.8$ & $39.4$ & $24.9$ & $49.1$ \\
		$2$ & $1$ & $66.0$ & $58.5$ & $48.9$ & $36.2$ & $20.8$ & $46.1$ \\
		\hline\hline
	\end{tabular}
	\caption{\textbf{Number of layers for the encoder and decoder.} 
	The table shows performances according to the number of layers for the encoder and decoder.}
	\label{tab:number_of_layers}
\end{table}
\endgroup

\begin{figure*}[t]
\centering
\includegraphics[width=17.40cm]{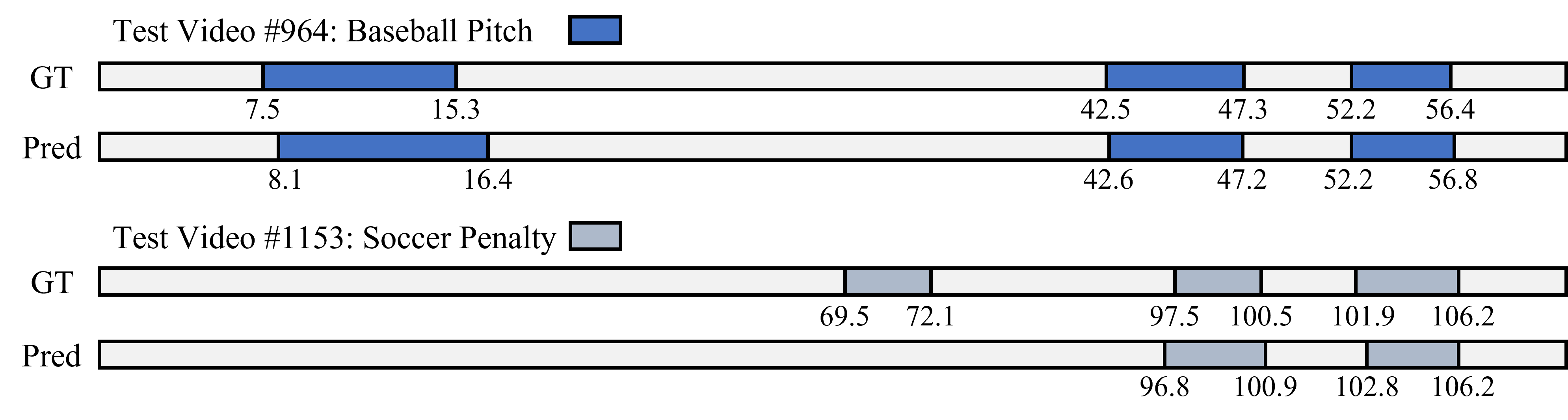}
\caption{\textbf{Qualitative results on THUMOS14.} The figure shows qualitative examples of Self-DETR for two validation videos in THUMOS14.
}
\label{fig:rebuttal_qualitative_results}
\end{figure*}

\begingroup
\setlength{\tabcolsep}{2.80pt} 
\renewcommand{\arraystretch}{1.0} 
\begin{table}[t]
	\centering
	\begin{tabular}{c|c|ccccc|c}
		\hline\hline
		Encoder & Dec.SA & $0.3$ & $0.4$ & $0.5$ & $0.6$ & $0.7$ & Avg. \\ \hline\hline
		$\cdot$ & $\cdot$ & $70.1$ & $62.7$ & $51.3$ & $36.5$ & $20.7$ & $48.3$ \\
        \checkmark & $\cdot$ & $\boldsymbol{70.7}$ & $\boldsymbol{64.7}$ & $\boldsymbol{54.0}$ & $38.7$ & $23.3$ & $\boldsymbol{50.3}$ \\
		$\cdot$ & \checkmark & $67.8$ & $61.8$ & $52.2$ & $\boldsymbol{40.8}$ & $\boldsymbol{24.6}$ & $49.4$ \\
		\checkmark & \checkmark & $70.5$ & $64.3$ & $53.9$ & $39.3$ & $23.8$ & $\boldsymbol{50.3}$ \\
		\hline\hline
	\end{tabular}
	\caption{\textbf{Ablation on collapsed self-attention.} 
	The table shows the results of ablation on collapsed self-attention of DETR without our self-feedback.}
	\label{tab:ablation_on_collpased_self_attention}
\end{table}
\endgroup

\begingroup
\setlength{\tabcolsep}{2.55pt} 
\renewcommand{\arraystretch}{1.00} 
\begin{table}[t]
	\centering
	\begin{tabular}{l|ccccc|c}
		\hline\hline
		Method & $0.3$ & $0.4$ & $0.5$ & $0.6$ & $0.7$ & Avg. \\
        \hline\hline
        DETR & $70.5$ & $64.3$ & $53.9$ & $39.3$ & $23.8$ & $50.3$ \\
        DETR + Self-feedback & $74.5$ & $69.5$ & $60.0$ & $47.6$ & $31.8$ & $56.7$ \\
        \hline
        DINO & $69.8$ & $63.1$ & $53.7$ & $41.5$ & $26.4$ & $50.9$ \\
        DINO + Self-feedback & $74.7$ & $69.4$ & $59.7$ & $46.8$ & $32.9$ & $56.7$   \\
		\hline\hline
	\end{tabular}
	\caption{\textbf{Recent DETR approach on THUMOS14}. The table shows the results on DINO~\cite{zhang2022dino}, a recent DETR method, with our self-feedback on THUMOS14.}
	\label{tab:rebuttal_recent_DETR}
\end{table}
\endgroup

\vspace{3pt}
\noindent\textbf{Recent DETR Approach with Self-Feedback.} 
Table.~\ref{tab:rebuttal_recent_DETR} shows the results of deploying a recent DETR approach, DINO~\cite{zhang2022dino}.
While DINO demonstrates excellent performance in object detection, simply deploying the denoising task does not enhance DETR for TAD.
Nevertheless, the self-feedback is still valid for DINO for TAD as temporal collapse persists with DINO.

\begin{figure}[t]
\centering
\includegraphics[width=8.35cm]{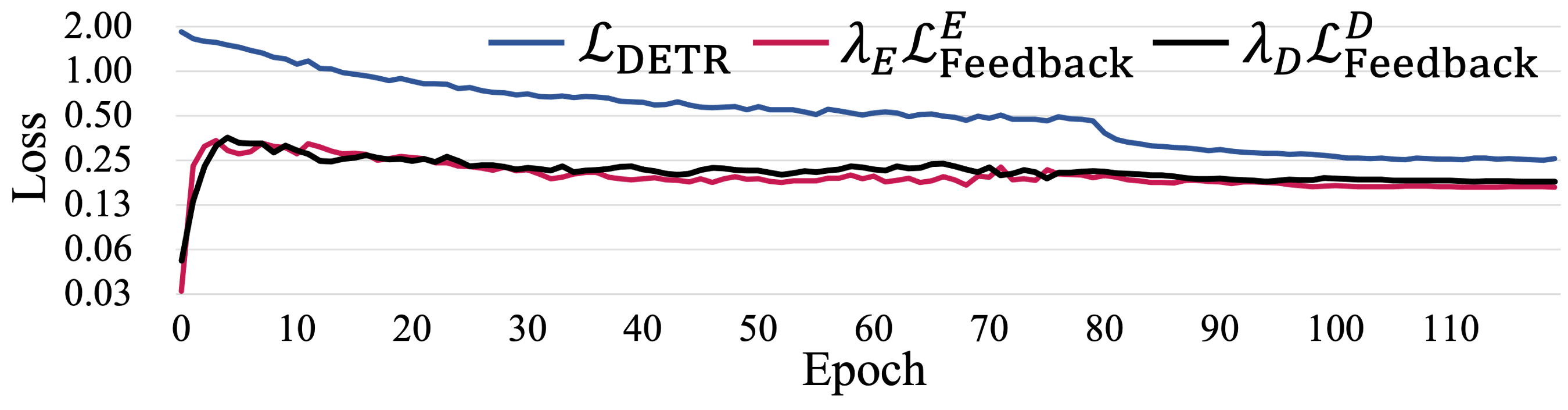}
\caption{\textbf{Losses of main and feedback objectives on THUMOS14.} The figure shows the training losses of main and feedback objectives over epochs on THUMOS14.
}
\label{fig:rebuttal_feedback}
\end{figure}

\begin{figure}[t]
\centering
\includegraphics[width=8.35cm]{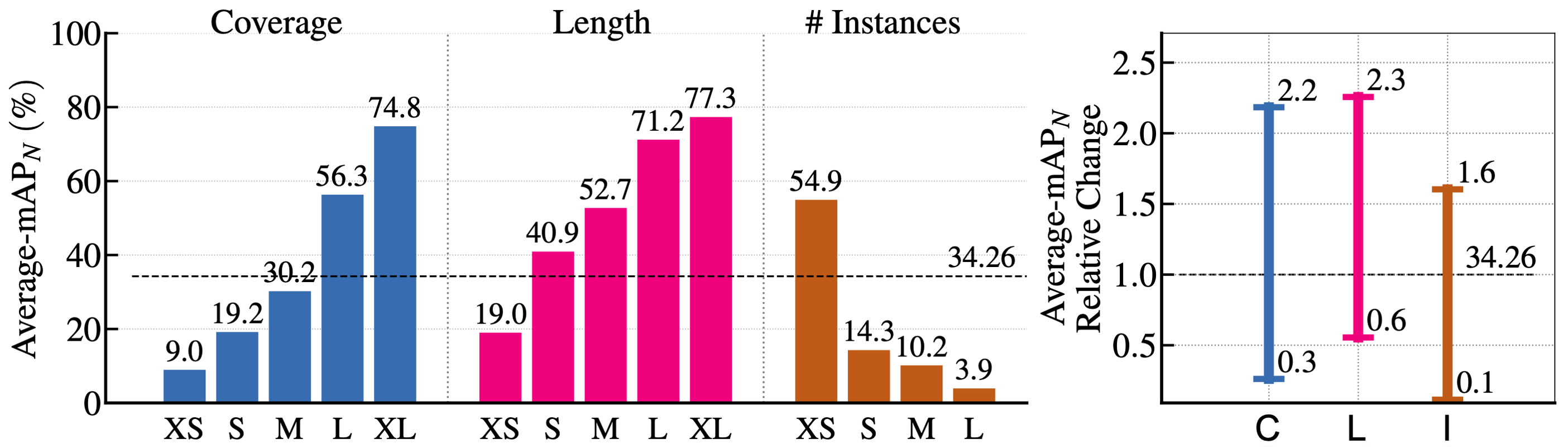}
\caption{\textbf{DETAD analysis on ActivityNet.} It shows the DETAD~\cite{alwassel2018detad} sensitivity analysis on ActivityNet.
}
\label{fig:rebuttal_sensitivity_analysis}
\end{figure}

\vspace{3pt}
\noindent\textbf{Self-feedback Losses.} 
We further analyzed the trend of the feedback losses according to the epoch as shown in Fig.~\ref{fig:rebuttal_feedback}.
Our proposed pipeline helps self-attention maps hold positions in the beginning and helps play their own roles finally through keeping the balance with the main objective.

\vspace{3pt}
\noindent\textbf{Error Analysis.} 
Fig.~\ref{fig:rebuttal_qualitative_results} illlustrates qualitative results for successful (upper) and failure (lower) cases.
Additionally, Fig.~\ref{fig:rebuttal_sensitivity_analysis} depicts sensitivity analysis of DETAD~\cite{alwassel2018detad} on ActivityNet.
In analysis, inferior performance of short scales is a crucial research concern for future work.

\vspace{3pt}
\noindent\textbf{Limitations.}
The DETR-based methods~\cite{tan2021relaxed, liu2021tadtr, shi2022react} including ours still lag behind in performance on ActivityNet compared to standard methods.
Unlike the THUMOS14 dataset, ActivityNet has much less instances in a video.
In this sense, the model is more likely to over-fit on long actions.
Therefore, imbalanced performance over scales is a crucial concern for the next step of DETR for TAD.

\end{document}